\documentclass{article}

\usepackage{microtype}
\usepackage{graphicx}
\usepackage{subcaption}
\usepackage{booktabs} 
\usepackage[table]{xcolor}

\usepackage{hyperref}

\usepackage[accepted]{icml2026}

\usepackage{amsmath}
\usepackage{amssymb}
\usepackage{mathtools}
\usepackage{amsthm}

\usepackage{xspace}
\usepackage{adjustbox}
\usepackage{wrapfig}
\usepackage{lipsum}
\usepackage{multirow}
\usepackage{makecell}
\usepackage{pifont}

\newcommand{\cmark}{\ding{51}}
\newcommand{\xmark}{\ding{55}}

\renewcommand{\algorithmicrequire}{\textbf{Input:}}
\renewcommand{\algorithmicensure}{\textbf{Output:}}

\newcommand{\name}[0]{TaH\xspace}
\newcommand{\xhdr}[1]{{\noindent\bfseries #1}.}

\newcommand{\rom}[1]{\textup{\uppercase\expandafter{\romannumeral#1}}}

\newcommand{\highlightblock}[1]{\begin{center}\vspace{-0.1cm}\emph{#1}\vspace{-0.1cm}\end{center}}

\newcounter{observation}

\newcounter{design}

\usepackage[most]{tcolorbox}
\usepackage{colortbl}
\usepackage{pifont}
\usepackage{makecell}

\tcbuselibrary{skins, breakable}   
\usepackage{tabularx, booktabs}    

\definecolor{darkred}{rgb}{0.5,0,0}
\definecolor{darkgreen}{rgb}{0,0.5,0}
\definecolor{tahgray}{RGB}{242,242,242}

\newtcolorbox[
  auto counter,
  list inside=examplelist,
]{greenbox}[2][]{
  enhanced,
  title={Example~\thetcbcounter. \textbf{#1}},
  colback=green!5!white,
  colframe=green!50!black,
  colbacktitle=green!60!black,
  coltitle=white,
  #2
}

\newtcolorbox[
  auto counter,
  list inside=examplelist
]{bluebox}[2][]{
  enhanced,
  title={Text~\thetcbcounter. \textbf{#1}},
  colback=cyan!5!white,  
  colframe=cyan!50!black,  
  colbacktitle=cyan!75!black,  
  coltitle=white,
  #2
}

\newcommand{\F}{\mathrm{FLOPs}}
\newcommand{\M}{\mathrm{Mem}}

\usepackage{amsmath,amsfonts,bm}

\def\eqref#1{equation~\ref{#1}}

\def\1{\bm{1}}

\DeclareMathAlphabet{\mathsfit}{\encodingdefault}{\sfdefault}{m}{sl}
\SetMathAlphabet{\mathsfit}{bold}{\encodingdefault}{\sfdefault}{bx}{n}

\newcommand{\softmax}{\mathrm{softmax}}

\usepackage[capitalize,noabbrev]{cleveref}

\theoremstyle{plain}

\theoremstyle{definition}

\theoremstyle{remark}

\usepackage[textsize=tiny]{todonotes}

\icmltitlerunning{Think-at-Hard: Dynamic Looped Transformers for Improved Reasoning}

\begin{document}

\twocolumn[
  \icmltitle{Think-at-Hard: \texorpdfstring{\\}{ }Dynamic Looped Transformers for Improved Reasoning}



  \icmlsetsymbol{equal}{*}

  \begin{icmlauthorlist}
    \icmlauthor{Tianyu Fu}{equal,tsinghua,infinigence}
    \icmlauthor{Yichen You}{equal,tsinghua}
    \icmlauthor{Zekai Chen}{tsinghua}
    \icmlauthor{Guohao Dai}{sjtu,infinigence}
    \icmlauthor{Huazhong Yang}{tsinghua}
    \icmlauthor{Yu Wang}{tsinghua}
  \end{icmlauthorlist}

  \icmlaffiliation{tsinghua}{Tsinghua University}
  \icmlaffiliation{infinigence}{Infinigence AI}
  \icmlaffiliation{sjtu}{Shanghai Jiao Tong University}

  \icmlcorrespondingauthor{Yu Wang}{yu-wang@tsinghua.edu.cn}

  \icmlkeywords{Machine Learning, ICML}

  \vskip 0.3in
]



\printAffiliationsAndNotice{\icmlEqualContribution}

\begin{abstract}
\label{sec:abstract}
Improving the reasoning abilities of Large Language Models (LLMs), especially under parameter constraints, is crucial for real-world applications.
Looped transformers address this by performing multiple latent iterations to refine each token beyond a single forward pass.
However, we identify a \emph{latent overthinking} phenomenon: most token predictions are already correct after the first pass, but are sometimes revised into errors in later iterations.
We ask whether \emph{selectively skipping} latent iterations can \emph{improve accuracy}, and reveal significant potential with an oracle iteration policy that boosts performance by up to 7.3\%.
Motivated by this, we propose Think-at-Hard (\name), a looped transformer optimized for selective iteration.
\name employs a lightweight neural decider to trigger latent iteration, only at tokens likely to be incorrect after the standard forward pass.
During latent iterations, depth-aware Low-Rank Adaptation (LoRA) modules shift the objective from general next-token prediction to focused hard-token refinement.
A duo-causal attention mechanism extends attention from the token sequence dimension to an additional iteration depth dimension, enabling cross-iteration information flow with full sequential parallelism.
Experiments on nine benchmarks show consistent gains across math, QA, and coding tasks.
With identical parameter counts, \name outperforms always-iterate baselines by 3.8-4.4\% while skipping iterations on 93\% of tokens, and exceeds single-iteration Qwen3 baselines by 3.0-3.8\%.
When allowing $<3\%$ more parameters from LoRA and decider, the gains further increase to 5.3-6.2\% and 6.1-6.8\%, respectively.
Our code is available \href{https://github.com/thu-nics/TaH}{here}.
\end{abstract}

\section{Introduction}
\label{sec:intro}

Recent advances in Large Language Model (LLM) reasoning have enabled broad applications across diverse domains~\citep{jaech2024openai-o1,guo2025deepseek-r1,yang2025qwen3}. 
With hundreds of billions of parameters, LLMs can generate complex Chain-of-Thought (CoT) to solve challenging tasks.
At the same time, smaller language models have also drawn increasing attention. 
With only a few billion parameters, they offer compelling alternatives: lower costs, faster inference, and suitability for edge computing~\citep{abdin2024phi3,minicpm4_2025,wang2025slm}.

At this crossroads, enhancing reasoning capabilities under parameter constraints becomes a central challenge. 
A common approach is to distill smaller models to mimic LLM CoT trajectories.
However, not all tokens are equally predictable: certain tokens encode critical logic or reasoning directions that are fundamentally harder to predict~\citep{lin2024rho, fu2025r2r, wang2025beyond}.
With limited computation per output token, small models quickly hit a performance ceiling and mispredict some of these tokens.
Once critical errors occur, the reasoning trajectory can irrecoverably diverge, yielding drastically different outcomes.

Prior work proposes looped transformers to address this parameter–performance paradox~\citep{hutchins2022block, saunshi2025loopedTrans, zeng2025pondering, zhu2025ouro}. 
Instead of verbalizing the next token immediately after one forward pass, these models typically feed the last-layer hidden states back into the LLM for additional passes, refining representations in the latent space.
After certain iterations, the final hidden states pass through the language modeling head to generate the next token.
By uniformly scaling up iterations per token, these models can correct initially wrong token predictions, potentially increasing performance without increasing parameter count.


However, we identify a \textit{latent overthinking} problem in looped transformers, where excessive iterations revise correct answers into wrong ones.
As shown in Figure~\ref{fig:landscape}, while the second iteration corrects 8.7\% of predictions, it also flips 2.1\% of correct ones into errors.
This occurs because most tokens, such as coherence or suffix tokens, are already predicted correctly after the first pass; further iterations may instead introduce harmful changes.
This mirrors overthinking in explicit CoT reasoning~\citep{wu2025morelesscot}, where additional reasoning steps degrade rather than improve answers.
This reveals a surprising opportunity:
\highlightblock{selectively skipping latent iterations on most tokens can further increase model performance.}
We validate this with an oracle policy that iterates only on initially mispredicted tokens, as shown in Table~\ref{tab:oracle_policy}. Compared to always iterating, this selective oracle can achieve up to 32\% higher accuracy with an optimized architecture.

Achieving selective latent iteration presents three main challenges. 
First, the model architecture should enable cross-depth attention, allowing each iteration to access full context.
This is crucial because when early tokens skip deeper iterations, later tokens must still access their representations from shallower depths.
Meanwhile, this cross-depth flow cannot compromise the sequence-level parallelism essential for efficient training and prefilling.
Second, the model must adapt to distribution shifts across iterations, while maximizing parameter reuse.
Third, training must remain stable despite tightly coupled dependencies: the iteration policy depends on prediction quality at each depth, while that quality depends on the depths to which previous tokens are assigned by the iteration policy.

\begin{figure}[t]
    \centering
    \includegraphics[width=\columnwidth]{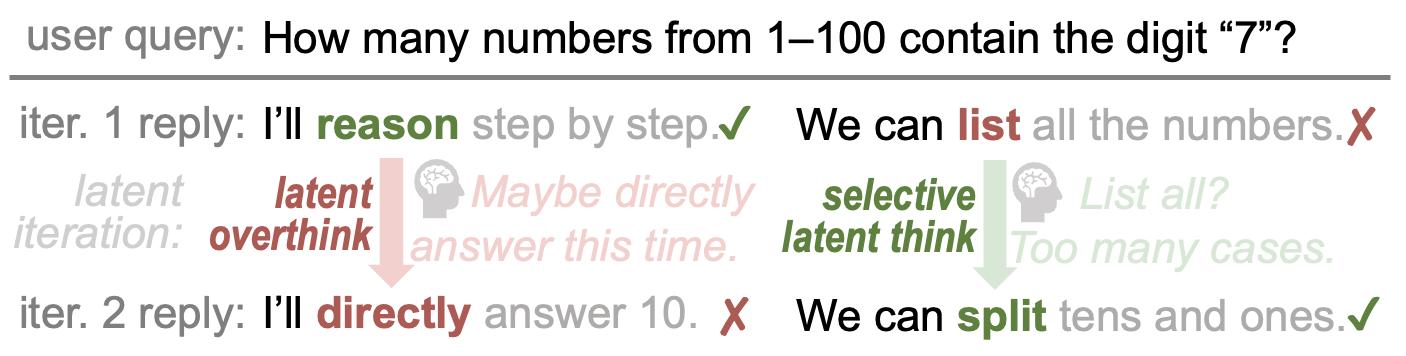}
    \caption{Latent iterations can fix wrong predictions, but can also \emph{overthink} and flip correct ones. \emph{Selective} iteration only when needed can improve reasoning with reduced computation.}   
    \label{fig:idea}
\end{figure}

To address these challenges, we propose \name, a looped transformer optimized for selective latent iteration.
As shown in Figure~\ref{fig:arch}, \name employs a neural decider to determine whether to iterate or verbalize each token.
We design a duo-causal attention mechanism to enable cross-depth attention with full sequence parallelism.
To specialize deeper iterations for current-token refinement and preserve strong first-pass predictions, we apply LoRA adapters solely at iterations $d>1$.
We train \name stably by aligning both the LLM backbone and iteration decider with a static oracle iteration policy.
We summarize our contributions as follows.

\begin{itemize}
    \item \xhdr{Selective Latent Iteration}
    We identify the latent overthinking phenomenon and quantify its influence on token prediction accuracy and downstream tasks.
    This insight motivates new directions where latent iteration is applied selectively to a few tokens for both better reasoning quality and efficiency.
    \item \xhdr{Specialized Model Architecture}
    We develop a model architecture that natively supports selective iterations.
    The dedicated duo-causal attention mechanism, LoRA adapters, and iteration decider enable efficient cross-depth information flow, objective transitions, and dynamic depth selection.
    \item \xhdr{Stable Training}
    We introduce a stable training scheme that uses a static oracle policy to decouple model adaptation and policy learning. It overcomes the circular dependency between iteration decisions and prediction quality.
\end{itemize}

We fine-tune \name from Qwen3-Base 0.6B and 1.7B on Open-R1, then test it across nine reasoning benchmarks spanning math, QA, and coding tasks. \name achieves average accuracy gains of 3.0\% and 3.8\% over standard single-iteration baselines.
Compared to looped transformer Ouro~\citep{zhu2025ouro} trained with same data, \name achieves 3.8-4.4\% gains while reducing latent iterations by 93\%.

\section{Related Work}
\label{sec:related_work}
Unlike standard LLMs that verbalize at every autoregressive step, latent thinking shifts part of generation away from explicit natural-language CoT in order to improve reasoning~\citep{li2025implicit}.

\xhdr{Signal-guided control}
These methods keep reasoning in token space but steer computation by inserting control tokens.
They add filler tokens (e.g., dots)~\citep{pfau2404let} and learnable \texttt{[PAUSE]} tokens~\citep{goyal2310think,kim2025learning} for extra compute during decoding.
They are lightweight but constrained to discrete-token interventions with limited latent control.

\xhdr{Latent optimization}
These methods perform autoregressive reasoning directly in internal representations, emitting little or no intermediate text.
They distill CoT into continuous embeddings via progressive replacement~\citep{hao2024training, cheng2024compressed}, hidden-state alignment~\citep{su2025token, liu2024expediting}, or logit-weighted embeddings~\citep{zhang2025soft}.
While efficient, these methods sacrifice interpretability, with training-based ones requiring heavy mitigation from verbal LLMs.

\xhdr{Looped transformers}
These methods interleave latent and verbal reasoning, adding latent iterations before each token verbalization.
Previous work focuses on scaling up iteration depths, with the main architectural focus on next-iteration inputs: reusing hidden states directly~\citep{saunshi2025loopedTrans,geiping2025scaling,zhu2025ouro} or using logit-weighted embeddings~\citep{zeng2025pondering}.
Looped transformers achieve deeper computation without parameter increases. However, uniform depth scaling burdens training and inference, and risks overthinking already-correct tokens.

\xhdr{Positioning}
\name belongs to the looped transformer family, but identifies \emph{selective iteration} as a new design principle to improve performance.
While concurrent works~\citep{bae2025mor, zhu2025ouro} also enable dynamic recursion, they degrade performance at non-maximum depths and require full retraining.
\name instead leverages existing pre-trained models, adding depth-aware LoRA and duo-causal attention to improve reasoning with minimal fine-tuning overhead.

\section{Preliminaries}
\label{sec:preliminary}

\xhdr{Autoregressive LLMs}
Modern LLMs generate text through an autoregressive next-token prediction process.
It includes a \emph{prefill} stage and a \emph{decode} stage~\citep{radford2018gpt1,radford2019gpt2,kwon2023vllm}.
In the prefill stage, the model processes the entire input sequence in parallel; in the decode stage, it consumes one new token at a time along with cached history to predict the next token.

Formally, let $t_i$ denote the token at position $i$ and $x_i\in\mathbb{R}^h$ its embedding. 
Let $E\in\mathbb{R}^{v\times h}$ be the embedding matrix, so $x_i=E[t_i]$ when $t_i$ is treated as an index. 
Here, $v$ and $h$ are the vocabulary size and hidden dimension. 
The output projection matrix is $W_{\text{out}}\in\mathbb{R}^{h\times v}$ (equal to $E^\top$ if tied). 
Given the context $T_{\le i}=[t_0,\dots,t_i]$ with embeddings $X_{\le i}=[x_0,\dots,x_i]$, the model $\theta$ produces a \emph{last-layer hidden state} $y_i$ for token $t_i$:
\begin{equation}
    y_i \;=\; \mathcal{P}_{\theta}\!\big(x_i \mid X_{\le i}\big)\;\in\;\mathbb{R}^h .
    \label{eq:llm_decode}
\end{equation}
The next-token distribution $p_i$ and sample are:
\begin{equation}
    p_i \;=\; \softmax\!\big(W_{\text{out}}^\top y_i\big) \in \mathbb{R}^v,
    \qquad
    t_{i+1} \;=\; \mathcal{S}(p_i),
    \label{eq:llm_sample}
\end{equation}
where $\mathcal{S}$ is a sampling rule such as nucleus sampling. 
Decoding repeats until an end-of-sequence token is generated.

\begin{figure*}[t]
    \centering
    \includegraphics[width=\textwidth]{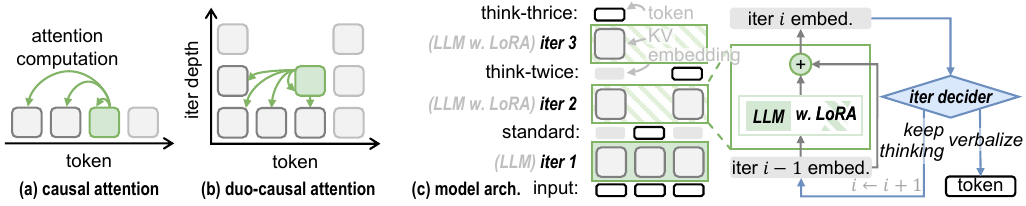}
    \caption{\name Overview. 
    (a) Regular causal attention: tokens attend only to previous positions.
    (b) Our duo-causal attention: tokens attend to both previous positions and shallower iteration depths, maintaining 2D causality.
    (c) Model architecture: \name selectively iterates or verbalizes tokens. It uses LoRA at deeper iterations to shift from next-token prediction to hard-token refinement. A neural decider determines whether to continue iterating or output the token.
    }
    \label{fig:arch}
\end{figure*}

\xhdr{Causal attention}
Modern LLMs typically adopt a \emph{causal} attention mechanism.
As shown in Figure~\ref{fig:arch}(a), each position attends only to itself and earlier positions, consistent with Equation~\ref{eq:llm_decode}. 
This design brings two key benefits: 
(1) it enables parallel training with next-token prediction and shifted logits, avoiding the need for token-by-token generation; 
and (2) it allows efficient inference by caching Key/Value states of past tokens instead of recomputing them.

\xhdr{Looped transformers}
Looped transformers introduce an inner loop that iterates in latent space before verbalizing each output token. 
Let $d\in\{1,2,\dots\}$ denote the iteration depth (written as a superscript), and set $x_i^{(0)}=E[t_i]$. 
At each iteration, looped transformers update $y_i$ with causal attention on the hidden states of \emph{the current iteration}:
\begin{equation}
    y_i^{(d)} \;=\; \mathcal{P}_{\theta}\!\big(x_i^{(d)} \,\big|\, X_{\le i}^{(d)}\big),
    \qquad
    X_{\le i}^{(d)}=[x_0^{(d)},\dots,x_i^{(d)}].
    \label{eq:rllm_decode}
\end{equation}
An inner transition then produces the next-depth embedding. 
For example, Loop~\citep{saunshi2025loopedTrans} simply sets $x_i^{(d+1)}=y_i^{(d)}$, while Ponder~\citep{zeng2025pondering} uses logit-weighted embeddings:
\begin{equation}
    x_i^{(d+1)}=\softmax\!\big(W_{\text{out}}^\top y_i^{(d)}\big)\ E = p_i^{(d)} E.
    \label{eq:ours_update}
\end{equation}
In practice, it uses the top-100 logits instead of full logits for efficiency.

Verbalization occurs at a fixed \emph{maximum depth} $d_{\max}$ shared by all tokens, where $y_i^{(d_{\max})}$ is transformed into the next token $t_{i+1}$, resembling Equation~\ref{eq:llm_sample}.

\section{Selective Latent Iteration Oracles}
\label{sec:oracle}

\begin{table}[t]
    \centering
    \footnotesize
    \setlength\tabcolsep{2pt}
    \caption{Performance comparison across iteration strategies, using Ouro-1.7B model except for the last row. Always1 means no latent iteration. Oracle policy iterates on 12-19\% of tokens. MMLU100 denotes the first 100 questions from MMLU-STEM.}
    \definecolor{tahgray}{RGB}{242,242,242}
    \begin{tabular}{l@{\hskip 3pt}cccc}
    \toprule
    \textbf{Iter. Policy} & \textbf{NTP} & \textbf{AMC23} & \textbf{MMLU100} & \textbf{HE++} \\
    \midrule
    Always1 & 73.1 & 38.1 & 56.0 & 39.6 \\
    Always2 & 79.7 & 40.6 & 60.0 & 40.9 \\
    \rowcolor{tahgray}
    Orcl.  &
    81.8{\scriptsize$/+2.1$} &
    47.9{\scriptsize$/+7.3$} &
    62.0{\scriptsize$/+2.0$}      &
    43.3{\scriptsize$/+2.4$} \\
    \midrule
    \rowcolor{tahgray}
    Orcl. w.\name &
    89.3{\scriptsize$/+9.6$} & 
    68.8{\scriptsize$/+28.2$} & 
    85.0{\scriptsize$/+25.0$} & 
    72.9{\scriptsize$/+32.0$}
          \\
    \bottomrule
    \end{tabular}
    \label{tab:oracle_policy}
\end{table}

\xhdr{Setup}
We analyze the latent iteration behavior of the Ouro looped transformer~\citep{zhu2025ouro} and our proposed \name model (architecture detailed in Section~\ref{sec:method}). 
All models are fine-tuned from Qwen3-1.7B-Base on a balanced 100K-sample subset of the Open-R1 dataset~\citep{openr1}, following the setups in Section~\ref{sec:eval_setup}.

\xhdr{Oracle iteration policy}
To investigate the potential of selective iteration, we establish an oracle policy $\pi$. It triggers additional iterations only when the reference LLM $\theta$ mispredicts the target token at the first forward pass.
In this section, we set $\theta$ to the model under evaluation, so $\pi$ acts as a greedy, locally optimal policy.

Formally, let $\hat{p}_i$ denote the reference model's first-pass next-token distribution at position $i$, and $t_{i+1}$ the ground-truth token.
The oracle iteration depth $d_i^{\pi}$ is:
\begin{equation}
    d_i^{\pi} \;=\; 1+\mathcal{D}(\hat{p}_i^{(1)}, t_{i+1}),
    \label{eq:ours_oracle_depth}
\end{equation}
where $\mathcal{D}$ is a binary discrepancy metric.
We use top-1 mismatch here: $\mathcal{D} = \1[\hat{t}_{i+1} \neq t_{i+1}]$, with $\hat{t}_{i+1}=\arg\max_{t} \hat{p}_i^{(1)}$ ($\1[\cdot]$ denotes the indicator function).
For simplicity, we assume $d_{\max}=2$. The arbitrary-depth case is in Appendix~\ref{sec:appendix/oracle_policy}, and alternative discrepancy metrics are ablated in Table~\ref{tab:ablation}.

For brevity, we call a token \emph{easy} if the reference model correctly predicts it at the first pass ($d_i^{\pi}=1$), and \emph{hard} otherwise.
Prior work~\citep{fu2025r2r} shows that \emph{hard} tokens typically occupy only a small proportion (e.g., $7\%$) of all tokens.

\xhdr{Next-token prediction}
We evaluate next-token prediction (NTP) accuracy by comparing top-1 predictions against ground-truth tokens in the Open-R1 validation set.
As shown in Table~\ref{tab:oracle_policy}, always iterating twice improves NTP accuracy for Ouro by 6.6\%, but at the cost of doubled depth.
Surprisingly, the oracle policy, which skips iterations on 81-88\% of tokens, further improves accuracy by 2.1\%.
This gain comes from avoiding \emph{latent overthinking}: without selective iteration, the model can revise correct predictions to wrong ones at the second pass, as shown in Figure~\ref{fig:landscape}.

\xhdr{Downstream tasks}
We next examine whether NTP gains translate to downstream task performance.
Since ground-truth tokens are unavailable during generation, we use top-1 predictions from the stronger Qwen3-8B as proxy labels.
Table~\ref{tab:oracle_policy} shows that the oracle policy improves downstream performance by 2.0-7.3\%.

\xhdr{\name design objectives}
The oracle experiments reveal two key insights.
First, models have significant untapped potential when iteration depth is selected correctly.
Second, the oracle policy is effective enough to learn from, even though other globally optimal policies may exist.
These findings motivate the design objectives of \name: (1) its model architecture should natively support selective iteration depths; and (2) following the oracle policy, its training objective for deeper iterations is not to cover all tokens, but to selectively focus only on the few failing tokens.
We later validate that \name-1.7B can better utilize the oracle policy to achieve $>25\%$ improvement, surpassing even Qwen3-4B.
While the oracle requires ground-truth tokens unavailable at inference, approximating it via neural networks is a promising direction.

\section{\name Design}
\label{sec:method}

We expand the key motivations and designs of \name in this section, including the duo-causal attention mechanism for cross-depth information flow (Section~\ref{sec:duo_attention}), the depth-adaptive model architecture with LoRA adapters (Section~\ref{sec:model_architecture}), and a two-stage training scheme (Section~\ref{sec:training}).

\subsection{Duo-Causal Attention}
\label{sec:duo_attention}
\xhdr{Motivation}
In looped transformers with fixed depth, standard causal attention on the current iteration's KV states incorporates all context (Equation~\ref{eq:rllm_decode}). 
However, dynamic iteration depths create a challenge: tokens at deeper levels cannot access hidden states of previous tokens that verbalized at shallower depths.
This poses a dilemma between requiring up-to-date context from all previous tokens and maintaining parallel training where depth-$d$ computations cannot depend on uncomputed deeper states ($d'>d$).
Existing approaches compromise on one of these aspects. Some sacrifice parallelism by allowing attention to deeper iterations~\citep{hao2024training}; others preserve parallelism by restricting attention to only the initial iteration's KVs~\citep{bae2025mor}.
To resolve this dilemma, we introduce a simple yet effective mechanism to maximize cross-depth information flow while maintaining high parallelism.

\xhdr{Duo-causal attention mechanism}
As shown in Figure~\ref{fig:arch}(b), duo-causal attention extends \emph{causality} to two dimensions, letting tokens attend across both previous positions and shallower iteration depths.
Formally, we extend the accessible set from Equation~\ref{eq:rllm_decode} to
\begin{equation}
    X_{\le i}^{(\le d)} \;=\; \{\, x_j^{(k)} \;|\; j\le i,\; k\le d \,\}.
    \label{eq:duo_causal_attention}
\end{equation}
When all tokens iterate only once (as in standard transformers), this reduces to regular causal attention. 
The duo-causal design achieves both full parallel training and cross-depth information flow.
At depth $d$, all tokens compute their depth-$d$ representations simultaneously using \emph{only and all} information from depths $1$ through $d$.

\begin{figure}[tb]
    \centering
    \includegraphics[width=\linewidth]{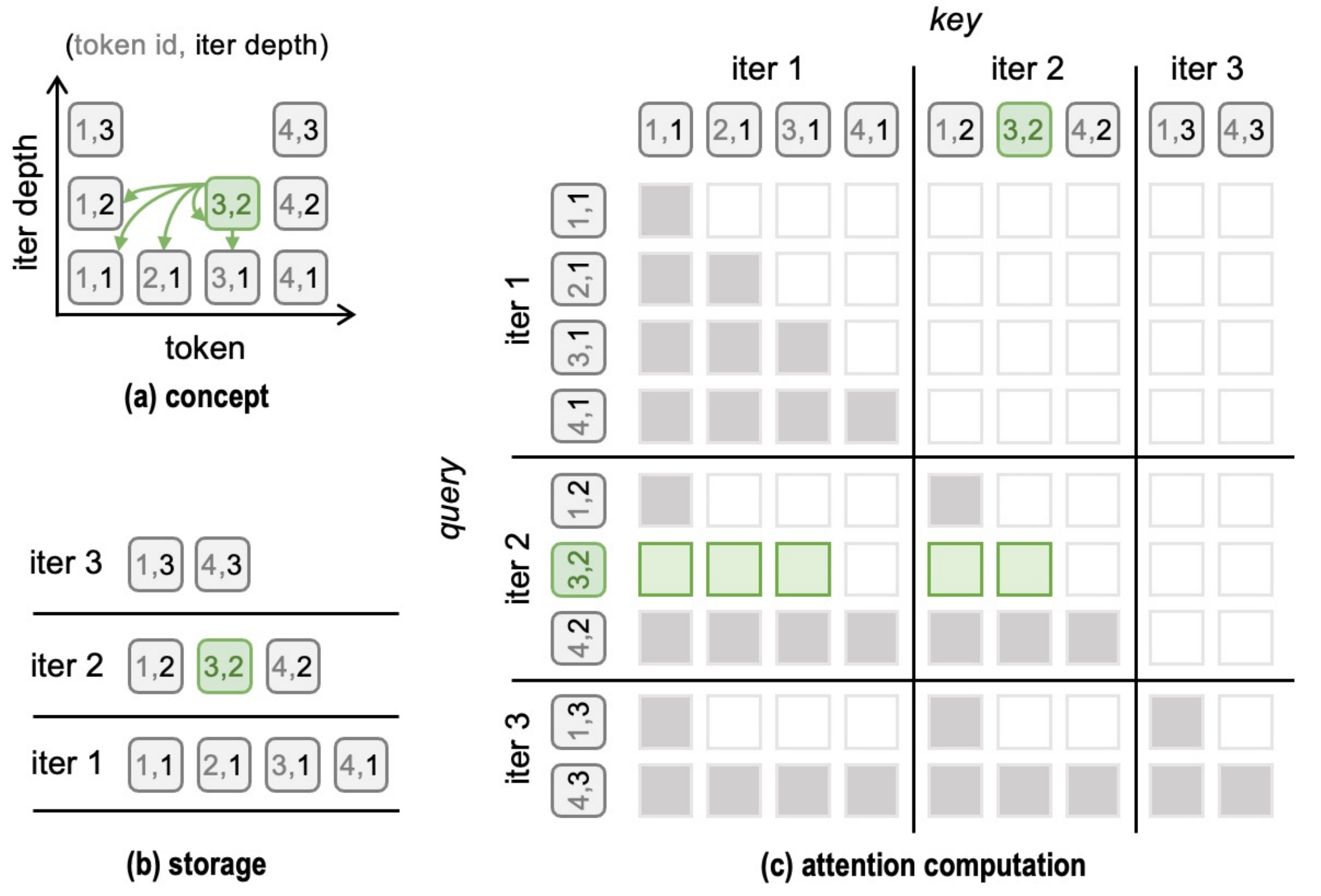}
    \caption{Duo-causal attention implementation. (a) Conceptual \name example with dynamic iteration depths. Each cell denotes a token–depth pair \((\text{token id}, \text{iter depth})\). (b) Each iteration maintains its own KV cache. (c) KV caches from all iterations are concatenated into a 1D sequence and processed with standard attention under a duo-causal mask. The duo-causal mask is conceptually partitioned into blocks by iteration depth. The diagonal blocks use a standard causal mask, while off-diagonal blocks use reduced causal masks that enforce the duo-causal rules.}
    \label{fig:duo_attention_expand}
\end{figure}

Duo-causal attention is fully compatible with attention kernels like FlashAttention~\citep{dao2022flashattention,dao2023flashattention2,shah2024flashattention3} and sparse implementations~\citep{fu2025moa, zhang2025spargeattention}.
As shown in Figure~\ref{fig:duo_attention_expand}, we simply maintain separate KV caches per iteration depth and flatten the 2D (token, depth) grid into 1D by concatenating deeper KV caches after shallower ones.
Positional encodings are applied based solely on the original token index, invariant to iteration depth.
The duo-causal constraint is then enforced via a modified additive attention mask, requiring no custom CUDA kernels.
More details on duo-causal attention implementation are discussed in Appendix~\ref{sec:appendix/duo-causal-attention}.


\subsection{Depth-adaptive Model Architecture}
\label{sec:model_architecture}

\xhdr{Motivation}
Previous looped transformers typically use identical weights across all iterations.
However, we find that over 73\% of next-tokens are correctly predicted at the first iteration (Table~\ref{tab:oracle_policy}).
This suggests deeper iterations serve a different objective: they refine the first iteration's prediction rather than predicting further ahead to the next-next token.
This mirrors deep LLMs, where shallow layers predict next tokens for deeper layers to refine~\citep{belrose2023tunedLens, schuster2022confidentEarlyExit, bae2023fastEarlyExit}.
While deep LLMs naturally handle this shift through distinct parameters per depth, looped transformers must accommodate both objectives with shared weights, potentially limiting performance.
Moreover, fixed iteration depths can cause \emph{latent overthinking}, motivating our dynamic approach.

\xhdr{Backbone model}
To address the objective shift, we apply a LoRA adapter~\citep{hu2022lora} to the shared LLM backbone only for iterations $d>1$.
As shown in Figure~\ref{fig:arch}(c), this allows the base LLM to focus on latent embeddings, while the adapter handles the objective shift.
It preserves strong next-token prediction at $d=1$, alleviating interference from deeper iterations.
We also add residual connections across iterations to simplify the refinement and improve gradient flow.
Formally, at depth $d$, we compute
\begin{equation}
    y_i^{(d)} \;=\; \mathcal{P}_{\theta_d}\!\Big(x_i^{(d)} \,\Big|\, X_{\le i}^{(\le d)}\Big),
    \label{eq:ours_decode}
\end{equation}
with depth-specific parameters
\[
\theta_d=\theta \ \text{for } d=1,
\qquad
\theta_d=\theta+\Delta \ \text{for } d > 1,
\]
where $\theta$ and $\Delta$ denote the LLM and LoRA weights, respectively.
The next-iteration inputs use logit-weighted embeddings (Equation~\ref{eq:ours_update}); verbalization follows standard sampling (Equation~\ref{eq:llm_sample}).
Each $y_i^{(d)}$ either continues iterating or verbalizes according to the decider $\mathcal{I}_\phi$.

\xhdr{Iteration decider}
We use a lightweight MLP as the iteration decider $\mathcal{I}_\phi$ to determine whether each token should continue iterating or verbalize.
After each iteration, it concatenates the backbone's shallow, middle, and final hidden states to predict a continuation probability:
\[
\hat{c}_i^{(d)} \;=\; \mathcal{I}_\phi\!\big(h_i^{(d)}\big) \in [0,1].
\]
During inference, token $i$ verbalizes when $\hat{c}_i^{(d)}$ falls below threshold $c_{\text{threshold}}$ or reaches maximum depth $d_{\max}$.

\subsection{Training Scheme}
\label{sec:training}

We adopt a two-stage training scheme. We first fine-tune the backbone model to support selective latent iteration, then train the iteration decider. Both stages are aligned to the same oracle iteration policy.

\xhdr{Motivation}
As shown in Figure~\ref{fig:arch}(c), the backbone LLM and the iteration decider are tightly coupled, making joint training unstable.
Specifically, the backbone’s prediction quality across iterations determines the optimal depth, while the decider controls the backbone’s KV cache and output depth. 
To stabilize training, we train the two components sequentially under a fixed oracle policy $\pi$ (validated in Section~\ref{sec:oracle}).

\paragraph{Oracle Iteration Policy $\pi$.}
Our goal is to iterate only on \emph{hard} tokens that a standard supervised fine-tuned (SFT) model would mispredict on the first pass.
Thus, we define $\pi$ with the SFT model as the reference: we trigger an additional iteration when its top-1 next-token prediction differs from the ground-truth token.
In principle, one could instead define $\pi$ using the current looped model itself (i.e., an on-policy oracle), but we find this to be empirically unstable (Section~\ref{sec:exp/dse}).


\paragraph{Stage 1: Backbone supervision under $\pi$.}
We optimize the backbone LLM ($\theta$ and LoRA adapter $\Delta$) with $\pi$-guided iteration execution.
The loss is standard next-token prediction at the oracle-determined depth:
\[
\mathcal{L}_{\text{SFT}}(\theta,\Delta)
\;=\;
\sum_i \; -\log p_i^{(d_i^{\pi})}\!\big(t_{i+1}\big),
\]
where $p_i^{(d_i^{\pi})}$ is the next-token distribution at position $i$, depth $d_i^{\pi}$.
This preserves first-iteration accuracy for easy tokens while training deeper iterations to refine hard tokens.

\paragraph{Stage 2: Decider imitation under frozen backbone.}
We freeze the backbone model $(\theta,\Delta)$ and train the iteration decider $\phi$ to imitate the oracle policy's continuation decisions.
We minimize weighted binary cross-entropy:
\[
\mathcal{L}_{\text{dec}}(\phi)
\;=\;
-\sum_{i,d} w_d 
c_i^{(d)} \log \hat{c}_i^{(d)} + (1-c_i^{(d)})\log(1-\hat{c}_i^{(d)}),
\]
where the sum ranges over tokens $i$ and depths $d=1,\ldots,\min\{d_{\max}-1, d_i^{\pi}\}$. Here $c_i^{(d)}$ is the ground-truth continuation label, $\hat{c}_i^{(d)}$ is the predicted probability, and $w_d$ is the class weight for label imbalance (ratio of stop to continue labels).



The two-stage scheme stabilizes training by decoupling backbone learning (conditioned on a fixed $\pi$) from iteration policy learning (imitation of $\pi$).

\section{Experiment}
\label{sec:experiment}

\subsection{Setup}

We present key experiment configurations here, with detailed setups in Appendix~\ref{sec:appendix/exp_setup}.

\xhdr{Baselines}
We compare diverse methods under equal parameter budgets, using Qwen3-\{0.6B,1.7B,4B\}-Base~\citep{yang2025qwen3} as backbones.
We compare \name over the following baselines:
(1) \emph{Standard}, which always verbalizes directly and reduces to the standard Qwen model;
(2) \emph{SoftThink}, a latent optimization method, implemented following the official design~\citep{zhang2025soft} on top of the Standard model;
(3) \emph{AlwaysThink}, similar to \name, but always iterates twice for all tokens during training and inference;
(4) \emph{Ouro}, a looped transformer that can scale iteration depths, implemented following the official design~\citep{zhu2025ouro}.
Unless otherwise specified, \name, Ouro, and AlwaysThink all use a maximum of two iterations, fine-tuned from the same Qwen3 backbone on the same training data and recipe.
Ouro reaches its highest performance when set to maximum iterations, so we report results under this setup.

\xhdr{\name setup}
To match the total parameter count of \name (with LoRA and decider) with that of baselines, we prune one layer from the LLM backbone before training.
The layer is chosen to minimize the increase in validation loss. 
We also report results for an unpruned variant, TaH+, which adds less than 3\% extra parameters from LoRA and decider. 
The detailed parameter composition is shown in Table~\ref{tab:param_composition}. 
Following~\citep{fu2025r2r}, we set the continuation threshold $c_{\text{threshold}}=0.9$ with about 7\% of tokens being iterated twice.
The oracle policy $\pi$ uses Qwen3-0.6B, 1.7B, and 4B as reference models, respectively.

\xhdr{Training scheme}
All models are trained on the balanced Open-R1~\citep{openr1} mixture (math, QA, and code; 100K samples) using supervised fine-tuning. 
To fit memory and compute limits, we exclude responses longer than 8,192 tokens; 4B models additionally truncate at 4,096 tokens; all other training settings follow the official Open-R1 script.
Each method is sufficiently trained for 5 epochs, and we select the checkpoint with the lowest validation loss as the final model. 
All backbones are initialized from the corresponding Qwen3-Base model.

\phantomsection
\label{sec:eval_setup}
\xhdr{Evaluation setup}
We evaluate across challenging reasoning benchmarks, including GSM8K~\citep{cobbe2021training}, MATH500~\citep{hendrycks2021measuring}, AMC23 (American
Mathematics Competitions), AIME25 (American Invitational Mathematics Examination), OlympiadBench (denoted as Olympiad)~\citep{he2024olympiadbench}, MBPP++~\citep{mbpp}, HumanEval++ (denoted as HE++)~\citep{humaneval}, GPQA-Diamond (denoted as GPQA)~\citep{rein2023gpqa}, and MMLU-STEM (denoted as MMLU)~\citep{mmlu}.
The maximum generation length is set to 8,192 tokens for all benchmarks, except GSM8K and MMLU-STEM, which use 4,096 due to their simpler problems and larger size.
Performance is reported as pass@1 under a zero-shot CoT setting, using a sampling temperature of 0.6.
We generate one sample per problem for large datasets (MATH500, OlympiadBench, etc.), and eight samples per problem for small datasets (AMC23, AIME25).

\subsection{Performance}
\label{sec:experiment/performance}

\xhdr{Benchmark evaluation}
We validate \name's reasoning ability across all nine benchmarks.
Table~\ref{tab:performance_by_category_transposed} presents performance results for models at 0.6B, 1.7B, and 4B parameter sizes.
Compared with the strong Standard Qwen3 baselines, we observe that existing approaches (AlwaysThink,
SoftThink, and Ouro) yield only marginal improvements when fine-tuned from base.
In contrast, \name achieves consistent gains across benchmarks.
For 0.6B and 1.7B models, \name delivers average improvements of 3.0\% and 3.8\% over Standard, respectively; TaH+, which adds less than 3\% additional parameters, further pushes these gains to 5.3\% and 6.2\%.
Compared to the concurrent work Ouro, \name and TaH+ achieve 3.8-4.4\% and 6.1-6.8\% gains, respectively.
For 4B models, which are trained with a 4K context length due to resource limits, \name and TaH+ also achieve gains of 1.7\% and 2.2\%.

\begin{table*}[t]
    \caption{Accuracy comparison across benchmarks.
    Best results are highlighted in bold. $^*$4B models are trained with $\leq$4K lengths due to resource constraints; AlwaysThink is excluded due to Out-Of-Memory (OOM) during training.}
    \label{tab:performance_by_category_transposed}
    \centering
    \setlength\tabcolsep{4pt} 
    \begin{tabular}{lcccccccccc}
    \toprule
    ~ &
    AIME25 & Olympiad & AMC23 & MATH500 & GSM8K & GPQA & MMLU & HE++ & MBPP++ & Average \\
    \midrule
    \multicolumn{11}{c}{\textit{0.6B}} \\
    \midrule
    Standard &
    1.9 & 15.4 & 22.7 & 39.9 & 58.2 & 31.1 & 54.2 & 16.8 & 28.8 & 29.9 \\
    SoftThink &
    \underline{2.9} & 14.0 & 22.2 & 39.6 & 55.9 & 24.7 & 53.0 & 14.3 & 29.5 & 28.5 \\
    Ouro &
    2.1 & 14.2 & 19.7 & 37.4 & 56.6 & \textbf{35.4} & 54.0 & 18.9 & 23.5 & 29.1 \\
    AlwaysThink &
    1.3 & 12.6 & 21.9 & 37.8 & 52.6 & 30.8 & 51.4 & 9.1 & 13.8 & 25.7 \\
    \rowcolor{tahgray}
    TaH &
    2.1 & \underline{19.1} & \underline{24.1} & \underline{46.2} &
    \underline{63.6} & 29.0 & \underline{56.4} & \underline{21.6} & \underline{33.9} & \underline{32.9}{\scriptsize$/+3.0$} \\
    \rowcolor{tahgray}
    TaH+ &
    \textbf{4.6} & \textbf{20.6} & \textbf{24.7} & \textbf{51.8} &
    \textbf{67.6} & \underline{31.3} & \textbf{59.0} & \textbf{22.0} & \textbf{35.1} & \textbf{35.2}{\scriptsize$/+5.3$} \\
    \midrule
    \multicolumn{11}{c}{\textit{1.7B}} \\
    \midrule
    Standard &
    10.8 & 33.8 & 39.7 & 67.8 & 80.2 & 30.3 & 74.1 & 39.0 & 51.9 & 47.5 \\
    SoftThink &
    5.4 & 30.7 & 40.3 & 64.8 & 80.0 & \underline{33.3} & 73.5 & 43.3 & 49.1 & 46.7 \\
    Ouro &
    10.8 & 31.3 & 40.6 & 68.2 & 79.8 & 32.8 & 72.4 & 40.9 & 45.5 & 46.9 \\
    AlwaysThink &
    7.5 & 31.0 & 40.9 & 63.2 & 74.2 & 30.5 & 69.6 & 16.4 & 25.6 & 39.9 \\
    \rowcolor{tahgray}
    TaH &
    \underline{13.8} & \underline{37.2} & \underline{40.9} &
    \underline{71.4} & \textbf{84.8} & \underline{33.3} &
    \underline{74.8} & \underline{50.0} & \underline{55.3} & \underline{51.3}{\scriptsize$/+3.8$} \\
    \rowcolor{tahgray}
    TaH+ &
    \textbf{15.4} & \textbf{37.6} & \textbf{48.4} &
    \textbf{72.6} & \underline{84.5} & \textbf{39.4} &
    \textbf{76.6} & \textbf{51.5} & \textbf{57.5} & \textbf{53.7}{\scriptsize$/+6.2$} \\
    \midrule
    \multicolumn{11}{c}{\textit{4B$^*$}} \\
    \midrule
    Standard &
    24.2 & 50.4 & 64.2 & 84.2 & \underline{91.7} &
    46.2 & 85.4 & 69.2 & 65.9 & 64.6 \\
    SoftThink &
    25.8 & 50.2 & 63.1 & 85.0 & \textbf{92.5} & \textbf{50.3} & 86.0 & 69.2 & 66.7 & 65.4 \\
    Ouro &
    25.0 & \underline{51.4} & 64.4 & 83.8 & 90.7 & \underline{50.0} & 85.9 & \underline{70.7} & 66.7 & 65.4 \\
    \rowcolor{tahgray}
    TaH &
    \underline{27.1} & 50.5 & \textbf{69.7} & \textbf{85.8} & 91.0 & 48.5
     & \underline{86.2} & 70.1
     & \underline{67.7} & \underline{66.3}{\scriptsize$/+1.7$}\\
    \rowcolor{tahgray}
    TaH+ &
    \textbf{27.9} & \textbf{52.6} & \underline{68.1} & \underline{85.6} & \underline{91.7} &
    49.0 & \textbf{86.6} & \textbf{72.0} & \textbf{68.1} & \textbf{66.8}{\scriptsize$/+2.2$} \\
    \bottomrule
    \end{tabular}
\end{table*}

\xhdr{Hardware-agnostic efficiency}
Tables~\ref{tab:token_iter_stats} and~\ref{tab:efficiency_analysis_theo} report the average iteration depth, per-token FLOPs, and memory access of \name.
On average, \name performs 1.07 iterations per token. It significantly undercuts the 2.08-2.18$\times$ FLOPs and memory access of AlwaysThink, matching the overhead of Standard with only 4-5\% additional overhead. See Appendix~\ref{sec:appendix/exp/theo_eff} for more details.

\begin{table}[tb]
    \caption{Real-world decoding performance on a single A800 GPU, tested on AIME25 with 8K max token length.}
    \label{tab:real_world_perf}
    \centering
    \setlength\tabcolsep{4pt}
    \begin{tabular}{lcc>{\columncolor{tahgray}}c}
    \toprule
     & Standard & AlwaysThink & \name \\
    \midrule
    Avg. Depth & 1.00 & 2.00 & 1.06 \\
    Memory (GB)        & 4.3   & 6.8   & 4.6   \\
    Latency (s)        & 210.6 & 747.2 & 301.4 \\
    Throughput (token/s) & 38.9 & 11.0  & 27.2  \\
    \bottomrule
    \end{tabular}
\end{table}

\xhdr{Real-world efficiency}
We compare efficiency of 1.7B models at 8K length on a single NVIDIA A800 GPU.
As shown in Table~\ref{tab:real_world_perf}, \name iterates twice on only 6\% of tokens on AIME25, with 1.48$\times$ lower memory overhead and 2.48$\times$ faster decoding than AlwaysThink, while achieving higher accuracy.
More detailed efficiency experiment setup and runtime breakdown are shown in Appendix~\ref{sec:appendix/exp/real_world_efficiency}.
    


\begin{figure}[tb]
    \centering
    \includegraphics[width=0.8\columnwidth]{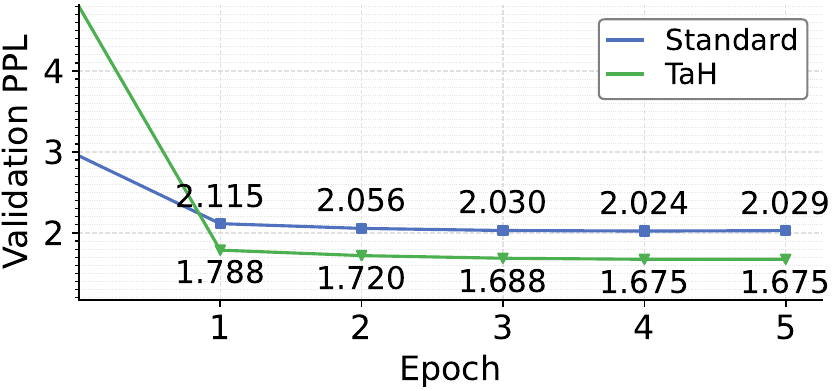}
    \caption{Training dynamics of the LLM backbone on Qwen3-0.6B-Base. \name converges rapidly and achieves lower perplexity.}
    \label{fig:training_llm}
\end{figure}

\begin{figure}[tb]
    \centering
    \includegraphics[width=0.8\columnwidth]{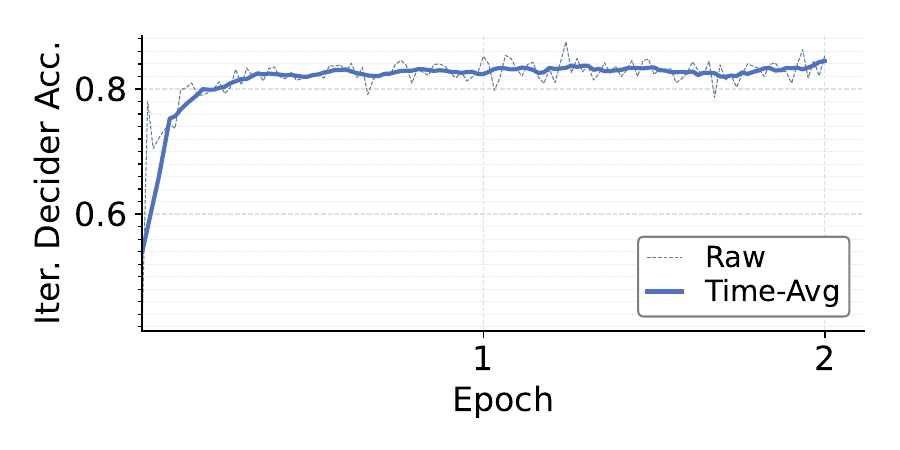}
    \caption{Iteration-decider accuracy vs. epoch (Qwen3-0.6B).}
    \label{fig:training_decider}
\end{figure}

\begin{figure}[tb]
    \centering
    \includegraphics[width=0.8\columnwidth]{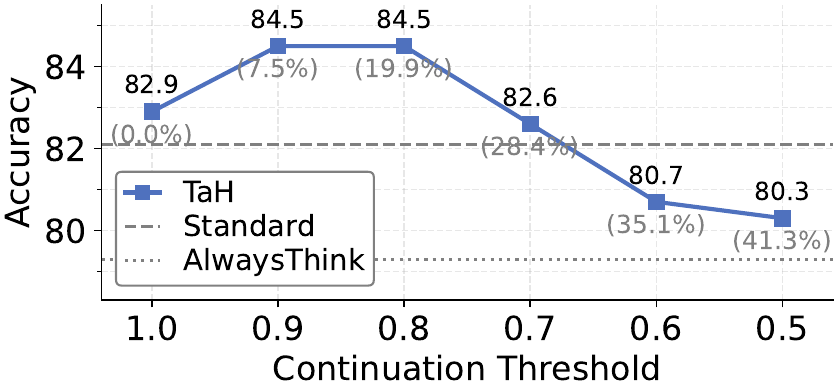}
    \caption{GSM8K accuracy with respect to continuation threshold. Numbers in brackets show the percentage of tokens iterated twice.}
    \label{fig:continuation_threshold}
\end{figure}

\xhdr{Training dynamics}
During stage~1 (LLM backbone training), \name performs iterations according to the oracle policy. As shown in Figure~\ref{fig:training_llm}, it converges notably faster than the Standard baseline and also achieves much lower validation perplexity.
During stage~2 (iteration-decider training), the neural decider successfully imitates the oracle strategy. It reaches about 83\% accuracy at predicting the oracle's iteration decisions, as shown in Figure~\ref{fig:training_decider}.

\subsection{Design Choice Exploration}
\label{sec:exp/dse}

We demonstrate the effectiveness and robustness of each design choice of \name through ablation studies.
All experiments in this section train \name and its variants on the math subset of Open-R1 and evaluate on MATH500, AMC23, and OlympiadBench.


\begin{table}[tb]
\caption{Ablation study on design choices of TaH-0.6B. Each row varies one aspect from the \name configuration (marked with \colorbox{tahgray}{gray}).}
\label{tab:ablation}
\centering
\footnotesize
\setlength\tabcolsep{2pt}
\begin{tabular}{l@{\hskip 0pt}cccc}
    \toprule
    \textbf{Variant} & \textbf{MATH500} & \textbf{AMC23} & \textbf{Olympiad} & \textbf{Average} \\
    \midrule
    \rowcolor{tahgray}
    \quad \name & \textbf{51.2} & \textbf{32.5} & \textbf{23.9} & \textbf{35.9} \\
    \midrule
    \multicolumn{5}{c}{\textit{Model Architecture}} \\
    \midrule
    \multicolumn{5}{l}{\textit{Iteration Depth} (\name: \colorbox{tahgray}{Neural decider})} \\
    \quad Always-1 & 47.2 & 23.4 & 18.8 & 29.8{\scriptsize$/-$6.1} \\
    \quad Always-2 & 32.8 & 15.6 & 10.2 & 19.5{\scriptsize$/-$16.4} \\
    \addlinespace
    \multicolumn{5}{l}{\textit{Attention} (\name: \colorbox{tahgray}{Duo-causal})} \\
    \quad Causal@iter1 & 47.8 & 24.4 & 19.4 & 30.5{\scriptsize$/-$5.4} \\
    \quad Causal@current & 42.0 & 23.8 & 16.4 & 27.4{\scriptsize$/-$8.5} \\
    \addlinespace
    \multicolumn{5}{l}{\textit{Depth Adapters} (\name: \colorbox{tahgray}{LoRA + Residual})} \\
    \quad w/o LoRA & 51.6 & 29.7 & 22.4 & 34.6{\scriptsize$/-$1.3} \\
    \quad w/o LoRA \& Res. & 49.2 & 22.5 & 21.2 & 31.0{\scriptsize$/-$4.9} \\
    \midrule
    \multicolumn{5}{c}{\textit{Training Scheme}} \\
    \midrule
    \multicolumn{5}{l}{\textit{Supervision Type} (\name: \colorbox{tahgray}{Token-only})} \\
    \quad Token+latent & 49.4 & 29.6 & 15.9 & 31.6{\scriptsize$/-$4.3} \\
    \addlinespace
    \multicolumn{5}{l}{\textit{Iteration Policy during LLM training} (\name: \colorbox{tahgray}{Oracle policy})} \\
    \quad Decider-based & 44.8 & 24.1 & 17.3 & 28.7{\scriptsize$/-$7.2} \\
    \quad Dynamic & 11.0 & 2.8 & 2.7 & 5.5{\scriptsize$/-$30.4} \\
    \addlinespace
    \multicolumn{5}{l}{\textit{Discrepancy metric in $\pi$} (\name: \colorbox{tahgray}{Top1 mismatch})} \\
    \quad Cross-entropy & 47.4 & 21.2 & 20.4 & 29.7{\scriptsize$/-$6.2} \\
    \quad Entropy & 42.0 & 21.9 & 16.9 & 26.9{\scriptsize$/-$9.0} \\
    \bottomrule
\end{tabular}
\end{table}

\xhdr{Model architecture}
(1) \textbf{Iteration Scheme}.
As shown in Table~\ref{tab:ablation}, decider-based iteration outperforms the \textit{Always-1} and \textit{Always-2} alternatives, confirming the practical benefits of selective iteration, even with imperfect decisions from the neural decider. Note that for \textit{Always-1}, duo-causal attention degenerates to regular causal attention.
(2) \textbf{Duo-Causal Attention}.
Replacing duo-causal attention with standard causal attention variants causes significant drops:
(a) attending only to the first iteration (Causal@iter1) drops by 5.4\%;
(b) attending only to the current iteration (Causal@current) drops even more, by 8.5\%.
They confirm duo-causal attention's essential role in cross-depth information flow.
(3) \textbf{Depth Adapters}.
Removing LoRA and residual connections leads to consistent drops, confirming their beneficial roles in objective transition across iterations.

\xhdr{Training scheme}
(1) \textbf{Supervision type.} 
Some previous work supervises all iteration depths with next-token labels, while \name only supervises final output tokens.
As shown in Table~\ref{tab:ablation}, such \textit{token+latent} supervision underperforms \name. This aligns with our intuition that different iterations should focus on their respective objectives.
(2) \textbf{Iteration policy during LLM training.}
We compare our static oracle strategy $\pi$ with two alternatives.
The \textit{decider-based} approach trains the iteration decider first, then uses it during backbone training. It suffers from the coupling challenge discussed in Section~\ref{sec:training}.
The \textit{dynamic} approach recalculates the oracle using the evolving backbone in Equation~\ref{eq:ours_oracle_depth}, facing the same coupling challenge and causing training collapse. 
These results support our selection of the static oracle policy.
(3) \textbf{Discrepancy metric in $\pi$.}
We compare three discrepancy metrics to trigger latent iteration in $\pi$: top-1 mismatch, entropy, and cross-entropy (detailed definitions in Appendix~\ref{sec:appendix/exp/oracle_label}).
As Table~\ref{tab:ablation} shows, top-1 mismatch yields the best result, confirming its empirical effectiveness.

\xhdr{Iteration label robustness}
We further evaluate how sensitive \name is to the quality of its oracle iteration labels.
For this evaluation, all \name-1.7B variants are trained on the math split while varying only how the iteration depth labels are constructed.
As shown in Table~\ref{tab:oracle_label_robustness}, labels from a smaller 0.6B reference and labels with 10\% random flips both still improve over Standard-1.7B.
The clean same-scale 1.7B reference gives the strongest average gain, so we use this label source in the main experiments.

\begin{table*}[t]
\caption{Robustness of \name-1.7B to imperfect oracle iteration labels. Each robustness row varies the label construction from the main \name configuration (marked with \colorbox{tahgray}{gray}); ``10\% noise'' randomly flips labels generated by the Qwen3-1.7B reference.}
\label{tab:oracle_label_robustness}
\centering
\footnotesize
\setlength{\tabcolsep}{6pt}
\begin{tabular}{llcccccc}
\toprule
\textbf{Method} & \textbf{Iter. Label} & \textbf{MATH500} & \textbf{GSM8K} & \textbf{AMC23} & \textbf{Olympiad} & \textbf{AIME25} & \textbf{Average} \\
\midrule
\multicolumn{8}{l}{\textit{Primary baselines}} \\
\quad Standard-1.7B & -- & 68.4 & 82.1 & 42.2 & 33.0 & 13.3 & 47.8 \\
\rowcolor{tahgray}
\quad \name-1.7B & 1.7B reference & \textbf{74.4} & \textbf{84.5} & \textbf{48.4} & \textbf{38.8} & \textbf{17.9} & \textbf{52.8}{\scriptsize$/+$5.0} \\
\addlinespace
\multicolumn{8}{l}{\textit{Label robustness checks}} \\
\quad \name-1.7B & 0.6B reference & 70.2 & 82.3 & 43.4 & 33.5 & 14.2 & 48.7{\scriptsize$/+$0.9} \\
\quad \name-1.7B & 1.7B reference + 10\% noise & \underline{73.6} & \underline{83.5} & \underline{46.8} & \underline{36.0} & \underline{14.6} & \underline{50.9}{\scriptsize$/+$3.1} \\
\bottomrule
\end{tabular}
\end{table*}

\begin{figure}[tb]
    \centering
    \includegraphics[width=0.8\columnwidth]{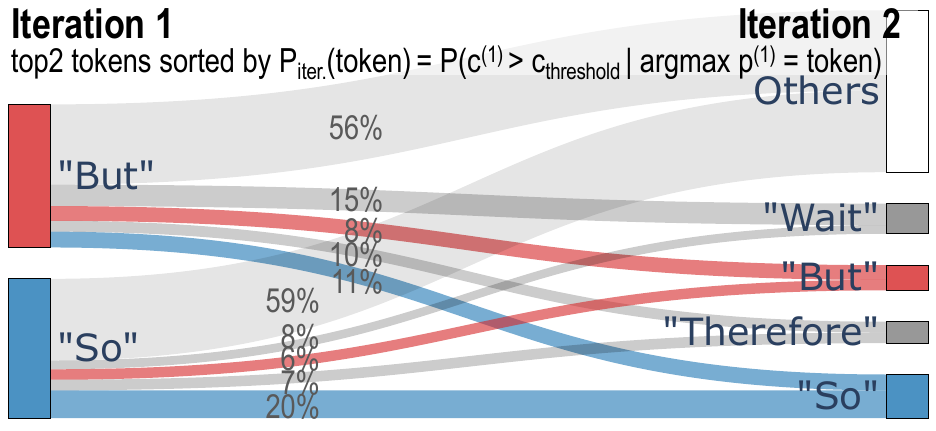}
    \caption{Next-token prediction changes across iterations. Top-2 frequent iterated tokens are visualized.}
    \label{fig:flow}
\end{figure}


\xhdr{Continuation threshold}
\name shows robust performance across continuation thresholds and iteration ratios (Figure~\ref{fig:continuation_threshold}).
We empirically set $c_{\text{threshold}}=0.9$ for all evaluations.

\begin{table*}[t]
\caption{Performance comparison with larger maximum iteration depths. TaH variants are all 1.7B, trained on the math subset. Brackets show the percentage of tokens executing iterations 2, 3, and 4.}
\label{tab:max_interation_3}
\centering
\small
\setlength{\tabcolsep}{2pt}
\newcommand{\iterpct}[1]{\hspace{0.35em}{\scriptsize[#1]}}
\begin{tabular*}{\textwidth}{@{\extracolsep{\fill}}llcccccc@{}}
\toprule
\textbf{Method} & \textbf{Iter. Policy} & \textbf{MATH500} & \textbf{GSM8K} & \textbf{AMC23} & \textbf{Olympiad} & \textbf{AIME25} & \textbf{Avg.} \\
\midrule
Standard & -- & 68.4\iterpct{0.0, 0.0, 0.0} & 82.1\iterpct{0.0, 0.0, 0.0} & 42.2\iterpct{0.0, 0.0, 0.0} & 33.0\iterpct{0.0, 0.0, 0.0} & 13.3\iterpct{0.0, 0.0, 0.0} & 47.8\iterpct{0.0, 0.0, 0.0} \\
\midrule
TaH-2 & \multirow{3}{*}{Decreasing} & \underline{74.4}\iterpct{5.6, 0.0, 0.0} & \underline{84.5}\iterpct{7.5, 0.0, 0.0} & 48.4\iterpct{4.2, 0.0, 0.0} & 38.8\iterpct{5.7, 0.0, 0.0} & 17.9\iterpct{6.0, 0.0, 0.0} & 52.8\iterpct{5.8, 0.0, 0.0} \\
TaH-3 & & \textbf{74.8}\iterpct{6.5, 1.2, 0.0} & 84.0\iterpct{8.6, 1.8, 0.0} & \underline{49.1}\iterpct{6.3, 1.0, 0.0} & \textbf{41.6}\iterpct{5.3, 1.0, 0.0} & \underline{19.6}\iterpct{5.4, 1.0, 0.0} & \underline{53.8}\iterpct{6.4, 1.2, 0.0} \\
TaH-4 & & \textbf{74.8}\iterpct{4.9, 0.6, 0.6} & \textbf{84.6}\iterpct{7.1, 0.9, 0.8} & \textbf{49.7}\iterpct{4.3, 0.5, 0.4} & \underline{40.7}\iterpct{4.0, 0.5, 0.4} & \textbf{20.4}\iterpct{4.3, 0.5, 0.3} & \textbf{54.0}\iterpct{4.9, 0.6, 0.5} \\
\midrule
TaH-3 & Uniform & 70.2\iterpct{29.4, 31.8, 0.0} & 82.5\iterpct{29.5, 35.8, 0.0} & 41.6\iterpct{30.7, 32.9, 0.0} & 34.4\iterpct{30.6, 34.1, 0.0} & 14.2\iterpct{30.4, 35.3, 0.0} & 48.6\iterpct{30.1, 34.0, 0.0} \\
\bottomrule
\end{tabular*}
\end{table*}

\xhdr{Deeper iteration depth}
We train 1.7B variants with larger maximum iteration depths on the math subset.
TaH-3 and TaH-4 set $D_{\max}=3$ and $D_{\max}=4$, respectively, following the same sparse, decreasing allocation principle as TaH-2: most tokens stop at depth~1, while harder tokens continue through additional latent passes.
As shown in Table~\ref{tab:max_interation_3}, the average gain over Standard increases from +5.0\% (TaH-2) to +6.0\% (TaH-3) and +6.2\% (TaH-4) with little extra cost.
We further study the impact of the label distribution by comparing with TaH-3-Uniform, which assigns same number of tokens across depths.
The weaker result of the uniform variant confirms that sparser allocation of deeper iterations is more effective than forcing frequent use of deeper depths.
Detailed setup is provided in Appendix~\ref{sec:appendix/oracle_policy}.

\xhdr{Restricted inference depth}
We evaluate whether the benefits of \name persist when the maximum inference depth is below the training maximum.
Table~\ref{tab:restricted_inference_depth} evaluates 1.7B models trained with $D_{\max}=2$ but forced to verbalize after the first iteration at inference.
Although weaker than full \name inference, restricted-depth inference still outperforms Standard by 2.1-3.1\% on average, suggesting that \name training also improves first-iteration representations.

\begin{table}[tb]
    \centering
    \caption{Performance of 1.7B models under restricted inference depth, grouped by task category.}
    \label{tab:restricted_inference_depth}
    \small
    \setlength{\tabcolsep}{3pt}
    \begin{tabular}{lllcccc}
    \toprule
    \textbf{Method} & \makecell{\textbf{Train.}\\\textbf{Depth}} & \makecell{\textbf{Infer.}\\\textbf{Depth}} & \textbf{Math} & \textbf{Science} & \textbf{Code} & \textbf{Avg.} \\
    \midrule
    Standard & 1 & 1 & 46.5 & 52.2 & 45.5 & 47.5 \\
    \midrule
    \multirow{2}{*}{\name} & 2 & 2 & 49.6 & \underline{54.1} & \underline{52.7} & \underline{51.3}{\scriptsize$/+$3.8} \\
     & 2 & 1 & 48.1 & 53.9 & 49.2 & 49.6{\scriptsize$/+$2.1} \\
    \midrule
    \multirow{2}{*}{TaH+} & 2 & 2 & \textbf{51.7} & \textbf{58.0} & \textbf{54.5} & \textbf{53.7}{\scriptsize$/+$6.2} \\
     & 2 & 1 & \underline{50.2} & 52.4 & 49.8 & 50.6{\scriptsize$/+$3.1} \\
    \bottomrule
    \end{tabular}
\end{table}


    
\subsection{Behavior Analysis}

\xhdr{Token alternation patterns}
We analyze which tokens \name selects for deeper iteration.
As shown in Figure~\ref{fig:flow}, \textit{But} and \textit{So} are iterated most frequently.
These tokens are hard to predict because they mark points of contrast or causality that can redirect subsequent reasoning.
At such junctures, \name uses additional iterations to refine its reasoning direction.
See Appendix~\ref{sec:appendix/exp/flow} for details.

\xhdr{Attention pattern}
We visualize the attention pattern of \name in Figure~\ref{fig:attention} and Appendix~\ref{sec:appendix/exp/attention}. Duo-causal attention focuses on different iterations in different heads, extracting broader contexts from multiple depths.

\begin{figure}[tb]
    \centering
    \includegraphics[width=0.67\columnwidth]{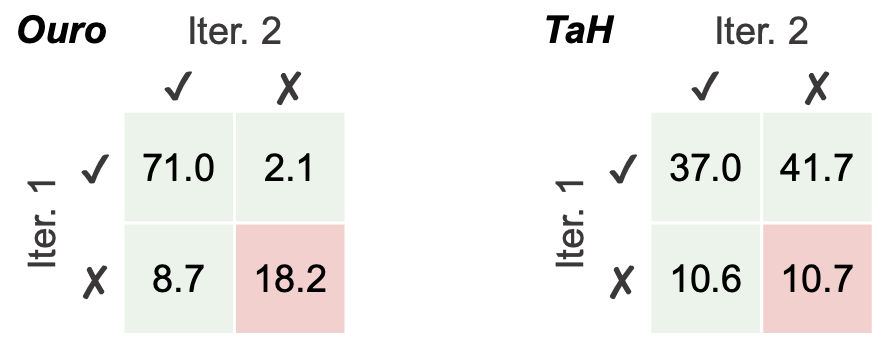}
    \caption{Distribution of token prediction accuracy across iterations for Ouro and \name.}
    \label{fig:landscape}
\end{figure}

\xhdr{Accuracy landscape}
Figure~\ref{fig:landscape} shows NTP accuracy across iterations. Because Ouro trains all iterations to predict all tokens, predictable tokens across depths largely overlap. 
\name specializes deeper iterations for hard tokens, improving overall coverage under selective iteration.

\section{Conclusion}
\label{sec:conclusion}
We present \name, a selective latent iteration method that simultaneously improves reasoning performance and efficiency.
\name introduces duo-causal attention, depth-specific LoRA, and a neural iteration decider, stably trained in two stages under a static oracle policy.
Across nine benchmarks, \name improves accuracy by 5.3-6.2\% over strong baselines with minimal overhead, opening a new direction for better reasoning under parameter constraints.

\clearpage
\section*{Acknowledgements}

This work was supported by National Natural Science Foundation of China (No. 62506197, 62325405, 62104128, U19B2019, U21B2031, 61832007, 62204164, 92364201), Tsinghua EE Xilinx AI Research Fund, and Beijing National Research Center for Information Science and Technology (BNRist).
We thank Xuefei Ning, and Donglin Yang for their valuable discussions and suggestions.
We also thank all the support from Infinigence-AI.
\section*{Impact Statement}

This paper presents work whose goal is to advance the field
of Looped Transformers. There are many potential societal
consequences of our work, none of which we feel must be
specifically highlighted here.

From a practical perspective, this paper improves model performance for parameter-constrained language models through selective iteration.
While this is an early exploration in this direction, we observe significant potential to simultaneously achieve accuracy gains comparable to much larger models while skipping iterations on most tokens.
By enhancing smaller models, our work may help democratize access to capable AI systems across diverse settings.
\bibliography{main}

@article{guo2025deepseek-r1,
  title={Deepseek-r1: Incentivizing reasoning capability in llms via reinforcement learning},
  author={Guo, Daya and Yang, Dejian and Zhang, Haowei and Song, Junxiao and Zhang, Ruoyu and Xu, Runxin and Zhu, Qihao and Ma, Shirong and Wang, Peiyi and Bi, Xiao and others},
  journal={arXiv preprint arXiv:2501.12948},
  year={2025}
}

@article{jaech2024openai-o1,
  title={Openai o1 system card},
  author={Jaech, Aaron and Kalai, Adam and Lerer, Adam and Richardson, Adam and El-Kishky, Ahmed and Low, Aiden and Helyar, Alec and Madry, Aleksander and Beutel, Alex and Carney, Alex and others},
  journal={arXiv preprint arXiv:2412.16720},
  year={2024}
}

@article{hao2024training,
  title={Training large language models to reason in a continuous latent space},
  author={Hao, Shibo and Sukhbaatar, Sainbayar and Su, DiJia and Li, Xian and Hu, Zhiting and Weston, Jason and Tian, Yuandong},
  journal={arXiv preprint arXiv:2412.06769},
  year={2024}
}

@article{cheng2024compressed,
  title={Compressed chain of thought: Efficient reasoning through dense representations},
  author={Cheng, Jeffrey and Van Durme, Benjamin},
  journal={arXiv preprint arXiv:2412.13171},
  year={2024}
}

@article{liu2024expediting,
  title={Expediting and elevating large language model reasoning via hidden chain-of-thought decoding},
  author={Liu, Tianqiao and Chen, Zui and Liu, Zitao and Tian, Mi and Luo, Weiqi},
  journal={arXiv preprint arXiv:2409.08561},
  year={2024}
}

@article{zhang2025soft,
  title={Soft thinking: Unlocking the reasoning potential of llms in continuous concept space},
  author={Zhang, Zhen and He, Xuehai and Yan, Weixiang and Shen, Ao and Zhao, Chenyang and Wang, Shuohang and Shen, Yelong and Wang, Xin Eric},
  journal={arXiv preprint arXiv:2505.15778},
  year={2025}
}

@article{su2025token,
  title={Token assorted: Mixing latent and text tokens for improved language model reasoning},
  author={Su, DiJia and Zhu, Hanlin and Xu, Yingchen and Jiao, Jiantao and Tian, Yuandong and Zheng, Qinqing},
  journal={arXiv preprint arXiv:2502.03275},
  year={2025}
}

@article{kim2025learning,
  title={Learning to Insert [PAUSE] Tokens for Better Reasoning},
  author={Kim, Eunki and Kim, Sangryul and Thorne, James},
  journal={arXiv preprint arXiv:2506.03616},
  year={2025}
}

@article{goyal2310think,
  title={Think before you speak: Training language models with pause tokens},
  year={2024},
  author={Goyal, Sachin and Ji, Ziwei and Rawat, Ankit Singh and Menon, Aditya Krishna and Kumar, Sanjiv and Nagarajan, Vaishnavh},
  journal={URL https://arxiv. org/abs/2310.02226}
}

@article{pfau2404let,
  title={Let’s think dot by dot: Hidden computation in transformer language models},
  year={2024},
  author={Pfau, Jacob and Merrill, William and Bowman, Samuel R},
  journal={URL https://arxiv. org/abs/2404.15758},
  volume={2404}
}

@article{raposo2024mixture,
  title={Mixture-of-depths: Dynamically allocating compute in transformer-based language models},
  author={Raposo, David and Ritter, Sam and Richards, Blake and Lillicrap, Timothy and Humphreys, Peter Conway and Santoro, Adam},
  journal={arXiv preprint arXiv:2404.02258},
  year={2024}
}

@article{luo2025adaptive,
  title={Adaptive layer-skipping in pre-trained llms},
  author={Luo, Xuan and Wang, Weizhi and Yan, Xifeng},
  journal={arXiv preprint arXiv:2503.23798},
  year={2025}
}

@article{del2023skipdecode,
  title={Skipdecode: Autoregressive skip decoding with batching and caching for efficient llm inference},
  author={Del Corro, Luciano and Del Giorno, Allie and Agarwal, Sahaj and Yu, Bin and Awadallah, Ahmed and Mukherjee, Subhabrata},
  journal={arXiv preprint arXiv:2307.02628},
  year={2023}
}

@article{li2025implicit,
  title={Implicit Reasoning in Large Language Models: A Comprehensive Survey},
  author={Li, Jindong and Fu, Yali and Fan, Li and Liu, Jiahong and Shu, Yao and Qin, Chengwei and Yang, Menglin and King, Irwin and Ying, Rex},
  journal={arXiv preprint arXiv:2509.02350},
  year={2025}
}

@article{bae2025mor,
  title={Mixture-of-recursions: Learning dynamic recursive depths for adaptive token-level computation},
  author={Bae, Sangmin and Kim, Yujin and Bayat, Reza and Kim, Sungnyun and Ha, Jiyoun and Schuster, Tal and Fisch, Adam and Harutyunyan, Hrayr and Ji, Ziwei and Courville, Aaron and others},
  journal={arXiv preprint arXiv:2507.10524},
  year={2025}
}

@article{fu2025r2r,
  title={R2R: Efficiently Navigating Divergent Reasoning Paths with Small-Large Model Token Routing},
  author={Fu, Tianyu and Ge, Yi and You, Yichen and Liu, Enshu and Yuan, Zhihang and Dai, Guohao and Yan, Shengen and Yang, Huazhong and Wang, Yu},
  journal={arXiv preprint arXiv:2505.21600},
  year={2025}
}

@article{yang2025qwen3,
  title={Qwen3 technical report},
  author={Yang, An and Li, Anfeng and Yang, Baosong and Zhang, Beichen and Hui, Binyuan and Zheng, Bo and Yu, Bowen and Gao, Chang and Huang, Chengen and Lv, Chenxu and others},
  journal={arXiv preprint arXiv:2505.09388},
  year={2025}
}

@misc{openr1,
    title = {Open R1: A fully open reproduction of DeepSeek-R1},
    url = {https://github.com/huggingface/open-r1},
    author = {{Hugging Face}},
    month = {January},
    year = {2025}
}

@article{cobbe2021training,
  title={Training verifiers to solve math word problems},
  author={Cobbe, Karl and Kosaraju, Vineet and Bavarian, Mohammad and Chen, Mark and Jun, Heewoo and Kaiser, Lukasz and Plappert, Matthias and Tworek, Jerry and Hilton, Jacob and Nakano, Reiichiro and others},
  journal={arXiv preprint arXiv:2110.14168},
  year={2021}
}

@article{hendrycks2021measuring,
  title={Measuring mathematical problem solving with the math dataset},
  author={Hendrycks, Dan and Burns, Collin and Kadavath, Saurav and Arora, Akul and Basart, Steven and Tang, Eric and Song, Dawn and Steinhardt, Jacob},
  journal={arXiv preprint arXiv:2103.03874},
  year={2021}
}

@article{he2024olympiadbench,
  title={Olympiadbench: A challenging benchmark for promoting agi with olympiad-level bilingual multimodal scientific problems},
  author={He, Chaoqun and Luo, Renjie and Bai, Yuzhuo and Hu, Shengding and Thai, Zhen Leng and Shen, Junhao and Hu, Jinyi and Han, Xu and Huang, Yujie and Zhang, Yuxiang and others},
  journal={arXiv preprint arXiv:2402.14008},
  year={2024}
}

@article{zeng2025pondering,
  title={Pretraining Language Models to Ponder in Continuous Space},
  author={Zeng, Boyi and Song, Shixiang and Huang, Siyuan and Wang, Yixuan and Li, He and He, Ziwei and Wang, Xinbing and Li, Zhiyu and Lin, Zhouhan},
  journal={arXiv preprint arXiv:2505.20674},
  year={2025}
}

@article{wu2025morelesscot,
  title={When more is less: Understanding chain-of-thought length in llms},
  author={Wu, Yuyang and Wang, Yifei and Ye, Ziyu and Du, Tianqi and Jegelka, Stefanie and Wang, Yisen},
  journal={arXiv preprint arXiv:2502.07266},
  year={2025}
}

@article{saunshi2025loopedTrans,
  title={Reasoning with latent thoughts: On the power of looped transformers},
  author={Saunshi, Nikunj and Dikkala, Nishanth and Li, Zhiyuan and Kumar, Sanjiv and Reddi, Sashank J},
  journal={arXiv preprint arXiv:2502.17416},
  year={2025}
}

@article{hutchins2022block,
  title={Block-recurrent transformers},
  author={Hutchins, DeLesley and Schlag, Imanol and Wu, Yuhuai and Dyer, Ethan and Neyshabur, Behnam},
  journal={Advances in neural information processing systems},
  volume={35},
  pages={33248--33261},
  year={2022}
}

@article{geiping2025scaling,
  title={Scaling up test-time compute with latent reasoning: A recurrent depth approach},
  author={Geiping, Jonas and McLeish, Sean and Jain, Neel and Kirchenbauer, John and Singh, Siddharth and Bartoldson, Brian R and Kailkhura, Bhavya and Bhatele, Abhinav and Goldstein, Tom},
  journal={arXiv preprint arXiv:2502.05171},
  year={2025}
}

@inproceedings{dao2022flashattention,
  title={Flash{A}ttention: Fast and Memory-Efficient Exact Attention with {IO}-Awareness},
  author={Dao, Tri and Fu, Daniel Y. and Ermon, Stefano and Rudra, Atri and R{\'e}, Christopher},
  booktitle={Advances in Neural Information Processing Systems (NeurIPS)},
  year={2022}
}

@inproceedings{dao2023flashattention2,
  title={Flash{A}ttention-2: Faster Attention with Better Parallelism and Work Partitioning},
  author={Dao, Tri},
  booktitle={International Conference on Learning Representations (ICLR)},
  year={2024}
}

@article{shah2024flashattention3,
  title={Flashattention-3: Fast and accurate attention with asynchrony and low-precision},
  author={Shah, Jay and Bikshandi, Ganesh and Zhang, Ying and Thakkar, Vijay and Ramani, Pradeep and Dao, Tri},
  journal={Advances in Neural Information Processing Systems},
  volume={37},
  pages={68658--68685},
  year={2024}
}

@article{belrose2023tunedLens,
  title={Eliciting latent predictions from transformers with the tuned lens},
  author={Belrose, Nora and Furman, Zach and Smith, Logan and Halawi, Danny and Ostrovsky, Igor and McKinney, Lev and Biderman, Stella and Steinhardt, Jacob},
  journal={arXiv preprint arXiv:2303.08112},
  year={2023}
}

@article{schuster2022confidentEarlyExit,
  title={Confident adaptive language modeling},
  author={Schuster, Tal and Fisch, Adam and Gupta, Jai and Dehghani, Mostafa and Bahri, Dara and Tran, Vinh and Tay, Yi and Metzler, Donald},
  journal={Advances in Neural Information Processing Systems},
  volume={35},
  pages={17456--17472},
  year={2022}
}

@article{bae2023fastEarlyExit,
  title={Fast and robust early-exiting framework for autoregressive language models with synchronized parallel decoding},
  author={Bae, Sangmin and Ko, Jongwoo and Song, Hwanjun and Yun, Se-Young},
  journal={arXiv preprint arXiv:2310.05424},
  year={2023}
}

@article{wang2025beyond,
  title={Beyond the 80/20 rule: High-entropy minority tokens drive effective reinforcement learning for llm reasoning},
  author={Wang, Shenzhi and Yu, Le and Gao, Chang and Zheng, Chujie and Liu, Shixuan and Lu, Rui and Dang, Kai and Chen, Xionghui and Yang, Jianxin and Zhang, Zhenru and others},
  journal={arXiv preprint arXiv:2506.01939},
  year={2025}
}

@article{lin2024rho,
  title={Rho-1: Not all tokens are what you need},
  author={Lin, Zhenghao and Gou, Zhibin and Gong, Yeyun and Liu, Xiao and Shen, Yelong and Xu, Ruochen and Lin, Chen and Yang, Yujiu and Jiao, Jian and Duan, Nan and others},
  journal={arXiv preprint arXiv:2404.07965},
  year={2024}
}

@article{chen2023eeLLM,
  title={Ee-llm: Large-scale training and inference of early-exit large language models with 3d parallelism},
  author={Chen, Yanxi and Pan, Xuchen and Li, Yaliang and Ding, Bolin and Zhou, Jingren},
  journal={arXiv preprint arXiv:2312.04916},
  year={2023}
}

@article{xie2023doremi,
  title={Doremi: Optimizing data mixtures speeds up language model pretraining},
  author={Xie, Sang Michael and Pham, Hieu and Dong, Xuanyi and Du, Nan and Liu, Hanxiao and Lu, Yifeng and Liang, Percy S and Le, Quoc V and Ma, Tengyu and Yu, Adams Wei},
  journal={Advances in Neural Information Processing Systems},
  volume={36},
  pages={69798--69818},
  year={2023}
}

@article{radford2018gpt1,
  title={Improving language understanding by generative pre-training},
  author={Radford, Alec and Narasimhan, Karthik and Salimans, Tim and Sutskever, Ilya and others},
  journal={OpenAI blog},
  year={2018},
  publisher={San Francisco, CA, USA}
}

@article{radford2019gpt2,
  title={Language models are unsupervised multitask learners},
  author={Radford, Alec and Wu, Jeffrey and Child, Rewon and Luan, David and Amodei, Dario and Sutskever, Ilya and others},
  journal={OpenAI blog},
  volume={1},
  number={8},
  pages={9},
  year={2019}
}

@inproceedings{kwon2023vllm,
  title={Efficient memory management for large language model serving with pagedattention},
  author={Kwon, Woosuk and Li, Zhuohan and Zhuang, Siyuan and Sheng, Ying and Zheng, Lianmin and Yu, Cody Hao and Gonzalez, Joseph and Zhang, Hao and Stoica, Ion},
  booktitle={Proceedings of the 29th symposium on operating systems principles},
  pages={611--626},
  year={2023}
}

@article{hu2022lora,
  title={Lora: Low-rank adaptation of large language models.},
  author={Hu, Edward J and Shen, Yelong and Wallis, Phillip and Allen-Zhu, Zeyuan and Li, Yuanzhi and Wang, Shean and Wang, Lu and Chen, Weizhu and others},
  journal={ICLR},
  volume={1},
  number={2},
  pages={3},
  year={2022}
}

@article{rein2023gpqa,
  title={Gpqa: A graduate-level google-proof q\&a benchmark},
  author={Rein, David and Hou, Betty Li and Stickland, Asa Cooper and Petty, Jackson and Pang, Richard Yuanzhe and Dirani, Julien and Michael, Julian and Bowman, Samuel R},
  journal={arXiv preprint arXiv:2311.12022},
  year={2023}
}

@article{abdin2024phi3,
  title        = {Phi-3 Technical Report: A Highly Capable Language Model Locally on Your Phone},
  author       = {Abdin, Marah and Aneja, Jyoti and Awadalla, Hany and Awadallah, Ahmed and Awan, Ammar Ahmad and Bach, Nguyen and Bahree, Amit and Bakhtiari, Arash and Bao, Jianmin and Behl, Harkirat and others},
  journal      = {arXiv preprint arXiv:2404.14219},
  year         = {2024},
  url          = {https://arxiv.org/abs/2404.14219}
}

@article{minicpm4_2025,
  title        = {MiniCPM4: Ultra‐Efficient LLMs on End Devices},
  author       = {MiniCPM Team and Chaojun Xiao and Yuxuan Li and Xu Han and Yuzhuo Bai and Jie Cai and Haotian Chen and Wentong Chen and Xin Cong and Ganqu Cui and Ning Ding and Shengdan Fan and Yewei Fang and Zixuan Fu and Wenyu Guan and Yitong Guan and Junshao Guo and Yufeng Han and Bingxiang He and Yuxiang Huang and Cunliang Kong and Qiuzuo Li and Zhen Li and Dan Liu and Biyuan Lin and Yankai Lin and Xiang Long and Quanyu Lu and Yaxi Lu and Peiyan Luo and Hongya Lyu and Litu Ou and Yinxu Pan and Zekai Qu and Qundong Shi and Zijun Song and Jiayuan Su and Zhou Su and Ao Sun and Xianghui Sun and Peijun Tang and Fangzheng Wang and Feng Wang and Shuo Wang and Yudong Wang and Yesai Wu and Zhenyu Xiao and Jie Xie and Zihao Xie and Yukun Yan and Jiarui Yuan and Kaihuo Zhang and Lei Zhang and Linyue Zhang and Xueren Zhang and Yudi Zhang and Hengyu Zhao and Weilin Zhao and Weilun Zhao and Yuanqian Zhao and Zhi Zheng and Ge Zhou and Jie Zhou and Wei Zhou and Zihan Zhou and Zixuan Zhou and Zhiyuan Liu and Guoyang Zeng and Chao Jia and Dahai Li and Maosong Sun},
  journal      = {arXiv preprint arXiv:2506.07900},
  year         = {2025},
  url          = {https://arxiv.org/abs/2506.07900}
}

@article{zhu2025ouro,
  title={Scaling Latent Reasoning via Looped Language Models},
  author={Zhu, Rui-Jie and Wang, Zixuan and Hua, Kai and Zhang, Tianyu and Li, Ziniu and Que, Haoran and Wei, Boyi and Wen, Zixin and Yin, Fan and Xing, He and others},
  journal={arXiv preprint arXiv:2510.25741},
  year={2025}
}

@article{zheng2024sglang,
  title={Sglang: Efficient execution of structured language model programs},
  author={Zheng, Lianmin and Yin, Liangsheng and Xie, Zhiqiang and Sun, Chuyue Livia and Huang, Jeff and Yu, Cody Hao and Cao, Shiyi and Kozyrakis, Christos and Stoica, Ion and Gonzalez, Joseph E and others},
  journal={Advances in neural information processing systems},
  volume={37},
  pages={62557--62583},
  year={2024}
}

@article{hoffmann2022training,
  title={Training compute-optimal large language models},
  author={Hoffmann, Jordan and Borgeaud, Sebastian and Mensch, Arthur and Buchatskaya, Elena and Cai, Trevor and Rutherford, Eliza and Casas, Diego de Las and Hendricks, Lisa Anne and Welbl, Johannes and Clark, Aidan and others},
  journal={arXiv preprint arXiv:2203.15556},
  year={2022}
}

@inproceedings{yang2024glitches,
  title={GLITCHES: GPU-FPGA LLM Inference Through a Collaborative Heterogeneous System},
  author={Yang, Fan and Yang, Xinhao and Wang, Hongyi and Wang, Zehao and Zhu, Zhenhua and Zeng, Shulin and Wang, Yu},
  booktitle={2024 IEEE High Performance Extreme Computing Conference (HPEC)},
  pages={1--7},
  year={2024},
  organization={IEEE}
}

@article{wenheng2025cdllm,
author = {Ma, Wenheng and Yang, Xinhao and Zeng, Shulin and Liu, Tengxuan and Shen, Libo and Wang, Hongyi and Li, Shiyao and Hong, Ke and Zhu, Zhenhua and Ning, Xuefei and Ho, Tsung-Yi and Dai, Guohao and Wang, Yu},
title = {CD-LLM: A Heterogeneous Multi-FPGA System for Batched Decoding of 70B+ LLMs using a Compute-Dedicated Architecture},
year = {2025},
publisher = {Association for Computing Machinery},
address = {New York, NY, USA},
issn = {1936-7406},
url = {https://doi.org/10.1145/3771288},
doi = {10.1145/3771288},
note = {Just Accepted},
journal = {ACM Trans. Reconfigurable Technol. Syst.},
month = oct
}

@article{wang2025slm,
  title={Slm-mux: Orchestrating small language models for reasoning},
  author={Wang, Chenyu and Wan, Zishen and Kang, Hao and Chen, Emma and Xie, Zhiqiang and Krishna, Tushar and Reddi, Vijay Janapa and Du, Yilun},
  journal={arXiv preprint arXiv:2510.05077},
  year={2025}
}

@inproceedings{fu2025moa,
  title={Mixture of Attention Spans: Optimizing LLM Inference Efficiency with Heterogeneous Sliding-Window Lengths},
  author={Fu, Tianyu and Huang, Haofeng and Ning, Xuefei and Zhang, Genghan and Chen, Boju and Wu, Tianqi and Wang, Hongyi and Huang, Zixiao and Li, Shiyao and Yan, Shengen and others},
  booktitle={Second Conference on Language Modeling},
  year=2025
}

@inproceedings{zhang2025spargeattention,
  title={SpargeAttention: Accurate and Training-free Sparse Attention Accelerating Any Model Inference},
  author={Zhang, Jintao and Xiang, Chendong and Huang, Haofeng and Wei, Jia and Xi, Haocheng and Zhu, Jun and Chen, Jianfei},
  booktitle={International Conference on Machine Learning (ICML 2025)},
  year={2025}
}

@article{mbpp,
  title={Program Synthesis with Large Language Models},
  author={Austin, Jacob and Odena, Augustus and Nye, Maxwell and Bosma, Maarten and Michalewski, Henryk and Dohan, David and Jiang, Ellen and Cai, Carrie and Terry, Michael and Le, Quoc and others},
  journal={arXiv preprint arXiv:2108.07732},
  year={2021}
}

@misc{humaneval,
      title={Evaluating Large Language Models Trained on Code},
      author={Mark Chen and Jerry Tworek and Heewoo Jun and Qiming Yuan and Henrique Ponde de Oliveira Pinto and Jared Kaplan and Harri Edwards and Yuri Burda and Nicholas Joseph and Greg Brockman and Alex Ray and Raul Puri and Gretchen Krueger and Michael Petrov and Heidy Khlaaf and Girish Sastry and Pamela Mishkin and Brooke Chan and Scott Gray and Nick Ryder and Mikhail Pavlov and Alethea Power and Lukasz Kaiser and Mohammad Bavarian and Clemens Winter and Philippe Tillet and Felipe Petroski Such and Dave Cummings and Matthias Plappert and Fotios Chantzis and Elizabeth Barnes and Ariel Herbert-Voss and William Hebgen Guss and Alex Nichol and Alex Paino and Nikolas Tezak and Jie Tang and Igor Babuschkin and Suchir Balaji and Shantanu Jain and William Saunders and Christopher Hesse and Andrew N. Carr and Jan Leike and Josh Achiam and Vedant Misra and Evan Morikawa and Alec Radford and Matthew Knight and Miles Brundage and Mira Murati and Katie Mayer and Peter Welinder and Bob McGrew and Dario Amodei and Sam McCandlish and Ilya Sutskever and Wojciech Zaremba},
      year={2021},
      eprint={2107.03374},
      archivePrefix={arXiv},
      primaryClass={cs.LG}
}

@article{mmlu,
      title={Measuring Massive Multitask Language Understanding},
      author={Dan Hendrycks and Collin Burns and Steven Basart and Andy Zou and Mantas Mazeika and Dawn Song and Jacob Steinhardt},
      journal={Proceedings of the International Conference on Learning Representations (ICLR)},
      year={2021}
    }
\bibliographystyle{icml2026}

\newpage
\appendix
\onecolumn

\section{Appendix}
\label{sec:appendix}





\subsection{Additional Experiment Setups}
\label{sec:appendix/exp_setup}

\subsubsection{Training Recipe}
\label{sec:appendix/exp_setup/training_recipe}

\begin{wraptable}{r}{0.40\linewidth}
\vspace{-\baselineskip}
\caption{Training hyperparameters.}
    \label{tab:train_hparams}
    \centering
    \begin{tabular}{ll}
    \toprule
    \textbf{Hyperparameter} & \textbf{Value} \\
    \midrule
    learning rate & 4e-5 \\
    max grad norm & 0.2 \\
    training epochs & 5 \\
    global batch size & 128 \\
    warmup ratio & 0.03 \\
    lr scheduler & cosine (min-lr ratio 0.1) \\
    precision & bfloat16 \\
    \bottomrule
    \end{tabular}
\end{wraptable}

We follow the official training setup of Open-R1~\citep{openr1} and use the Mixture-of-Thoughts dataset.
For our main performance experiments (Section~\ref{sec:experiment/performance}), we filter samples with output length exceeding 8K tokens from each category (math, code, and QA), then randomly sample 33K examples from each to form a balanced 100K training set.
The filtered dataset contains 480M tokens, with 1\% reserved for validation. 
For design choice exploration (Section~\ref{sec:exp/dse}), we use only the math subset filtered by 8K output length, resulting in 75K training samples.
For 0.6B and 1.7B models, we use a maximum sequence length of 8192 tokens for all methods.
For 4B models, we reduce the maximum sequence length to 4096 for Standard, \name, TaH+, and 3072 for Ouro due to memory constraints.
Detailed training hyperparameters are listed in Table~\ref{tab:train_hparams}.

\subsubsection{Baseline Setups}
\label{sec:appendix/setup/baseline}

\xhdr{AlwaysThink}
AlwaysThink uses the same architecture as TaH, except that it iterates twice at every token position and uses standard causal attention (attending only to the current iteration depth) instead of duo-causal attention.

\xhdr{Ouro}
We implement the Ouro architecture following its official design~\citep{zhu2025ouro}, but fine-tune it on the Open-R1 dataset from Qwen-Base initialization, to align with all other methods.
During fine-tuning, we adopt the entropy-regularized loss of Ouro.
Since the original Ouro is trained from scratch on different data distributions, performance may differ from the original paper.

\subsubsection{Iteration Depth Flexibility}
\label{sec:appendix/oracle_policy}

The main experiments in this paper focus on $D_{\max}=2$.
Here we detail how the oracle iteration policy generalizes to arbitrary maximum iteration depth, as used in Section~\ref{sec:exp/dse}.
The focus of this work is how to \emph{selectively attribute} iterations to different tokens, rather than increasing the average iteration depth.

\xhdr{Binary case ($D_{\max}=2$)}
Prior work has proposed many ways to estimate token difficulty, including excess loss~\citep{lin2024rho, xie2023doremi}, entropy~\citep{wang2025beyond, chen2023eeLLM}, and prediction difference~\citep{fu2025r2r}.
For shallow iteration budgets of up to two, we adopt the prediction-difference policy used in the main experiments: tokens that fail top-1 next-token prediction at the first iteration are labeled as hard.
Formally, the oracle iteration depth $d_i^\pi$ follows a binary rule:
\begin{equation}
    d_i^\pi \;=\; \begin{cases}
        1, & \text{if } h_i = 0 \quad \text{(easy token)} \\
        D_{\max}, & \text{if } h_i = 1 \quad \text{(hard token)} ,
    \end{cases}
\end{equation}
where $h_i$ is the hard-token indicator.
This induces a Bernoulli distribution over depths: easy tokens verbalize at depth~1, hard tokens iterate to depth~$D_{\max}$.

\xhdr{General framework ($D_{\max}>2$)}
For deeper iteration budgets, we replace the binary rule with a continuous difficulty score and a target depth distribution.
We use the reference model's first-pass cross-entropy as a continuous measure of token difficulty:
\[
\ell_i^{\text{ref}} \;=\; -\log p_{i,\text{ref}}^{(1)}\!\big(t_{i+1}\big).
\]
Let $\Lambda$ be a target distribution over $\{1,\dots,D_{\max}\}$ specifying the desired fraction of tokens at each depth.
We assign iteration depths via quantile mapping: rank all tokens by $\ell^{\text{ref}}$ to obtain $u_i = \mathrm{QuantileRank}(\ell_i^{\text{ref}}) \in [0,1]$, then set
\begin{equation}
    d_i^\pi \;=\; F_\Lambda^{-1}(u_i),
\end{equation}
where $F_\Lambda^{-1}$ is the quantile function (inverse CDF) of $\Lambda$.
Harder tokens (higher loss) thus receive deeper iterations, and the overall depth distribution follows $\Lambda$.
The binary case is recovered when $\Lambda$ matches the depth distribution induced by the hard-token labels (i.e., mass at depths~1 and~$D_{\max}$ equal to the easy and hard fractions, respectively).
The oracle depth $d_i^\pi$ is then converted into per-depth supervision for the iteration decider (Section~\ref{sec:training}): at each depth $d \in \{1,\ldots,D_{\max}-1\}$, the continuation label $c_i^{(d)} = \mathbb{1}[d < d_i^\pi]$ equals $1$ (continue iterating) when the token has not yet reached its assigned depth, and $0$ (verbalize) once $d = d_i^\pi$.
At inference, the decider's predicted continuation probabilities $\hat{c}_i^{(d)}$ are compared against the threshold $c_{\text{threshold}}$ to decide whether each token continues at each depth.

\xhdr{Deeper-depth experiment setup}
For TaH-3-Decreasing in Table~\ref{tab:max_interation_3}, we set the training target distribution over depths to $\Lambda=(0.88, 0.06, 0.06)$.
For TaH-4-Decreasing, we set $\Lambda=(0.88, 0.04, 0.04, 0.04)$.
That is, we keep the depth-1 fraction the same as TaH-2 and evenly allocate the remaining budget to deeper depths; the inference continuation threshold is set to $c_{\text{threshold}}=0.9$, which makes the realized continuation rate decrease with depth.
For comparison, TaH-3-Uniform uses $\Lambda=(0.34, 0.33, 0.33)$ during training and $c_{\text{threshold}}=0.5$ during inference to preserve a roughly uniform depth allocation.
The realized continuation ratios are reported in brackets in Table~\ref{tab:max_interation_3}.

\subsubsection{Parameter Breakdown}
\label{sec:appendix/exp_setup/parameter_breakdown}

Table~\ref{tab:param_composition} reports the parameter breakdown of the Standard, \name, and TaH+ methods.
To offset the additional parameters introduced by \name through LoRA and the iteration decider, we remove one layer from the LLM backbone, ensuring a fair comparison.
In practical deployments, we recommend TaH+, which adds only about 3\% additional parameters.

\begin{table}[b]
    \caption{Parameter breakdown of Standard, TaH, and TaH+. Counts are reported using M (million) and B (billion).}
    \label{tab:param_composition}
    \centering
    \begin{tabular}{ll|cccc}
    \toprule
    \textbf{Param.} & \textbf{Method} & \textbf{Backbone} & \textbf{LoRA} & \textbf{Iter. Decider} & \textbf{Total} \\
    \midrule
    \multirow{3}{*}{\textit{0.6B}} & Standard & 596M & -- & -- & 596M \\
    & \name & 580M & 10M & 5M & 595M \\
    & TaH+ & 596M & 10M & 5M & 611M \\
    \midrule
    \multirow{3}{*}{\textit{1.7B}} & Standard & 1.72B & -- & -- & 1.72B \\
    & \name & 1.67B & 17M & 18M & 1.71B \\
    & TaH+ & 1.72B & 17M & 18M & 1.76B \\
    \midrule
    \multirow{3}{*}{\textit{4B}} & Standard & 4.02B & -- & -- & 4.02B \\
    & \name & 3.92B & 32M & 70M & 4.02B \\
    & TaH+ & 4.02B & 33M & 70M & 4.12B \\
    \bottomrule
    \end{tabular}
\end{table}



\subsection{Additional Experimental Results}

\subsubsection{Math-Only Training}
This section isolates the math-only setting: models are trained on the math subset of Open-R1 and evaluated on math benchmarks as well as out-of-domain tasks.


\begin{table*}[t]
    \caption{Accuracy comparison of different baselines across five benchmarks and three model sizes.
    Subscripts indicate improvement over Standard. The top two scores for each task and model size are highlighted in bold.
    }
    \label{tab:math_performance}
    \centering
    \setlength\tabcolsep{3pt} 
    \begin{tabular}{ll|cccccc}
    \toprule
    & & \multicolumn{6}{c}{\textbf{Method}} \\
    \cmidrule(lr){3-8}
    \textbf{Param.} & \textbf{Benchmark} & \textbf{Standard} & \textbf{Routing}& \textbf{SoftThink}& \textbf{AlwaysThink}& \textbf{\name} & \textbf{TaH+} \\
    \midrule
    \multirow{7}{*}{\textit{0.6B}}
    & AIME25 & \textbf{4.2} & 1.0 & 2.5 & 1.5 & \textbf{4.2} & \textbf{5.0} \\
    & OlympiadBench & 18.8 & 7.4  & 19.4 & 10.2 & \textbf{23.9}  & \textbf{24.0} \\
    & AMC23 & 23.4 & 10.9 & 24.1 & 15.6 & \textbf{32.5} &  \textbf{30.6} \\
    & MATH500 & 47.2 & 27.3 & 48.8 & 32.8 & \textbf{51.2} & \textbf{54.2} \\
    & GSM8K & 62.5 & 45.6  & 61.3 & 54.6 & \textbf{64.4} & \textbf{68.8} \\
    \cmidrule(lr){2-8}
    & Average & 31.2 & 18.5 & 31.2 & 22.9 & \textbf{35.2{\scriptsize$/+$4.0}} & \textbf{36.5{\scriptsize$/+$5.3}} \\

    \midrule
    \multirow{7}{*}{\textit{1.7B}}& AIME25 & 13.3 & 10.2 & 12.9 & 10.0 & \textbf{17.9} & \textbf{14.6} \\
    & OlympiadBench & 33.0 & 30.6 & 33.4 & 30.0 & \textbf{38.8} & \textbf{41.2} \\
    & AMC23 & 42.2 & 42.2 & 43.1 & 42.5 & \textbf{48.4} & \textbf{51.2} \\
    & MATH500 & 68.4 & 60.0 & 68.8 & 61.8 & \textbf{74.4} & \textbf{73.0} \\
    & GSM8K & 82.1 & 71.2 & 79.6 & 79.3 & \textbf{84.5} & \textbf{85.8} \\
    \cmidrule(lr){2-8}
    & Average & 47.8  & 36.8 & 47.6 & 44.7 & \textbf{52.8{\scriptsize$/+$5.0}}& \textbf{53.2{\scriptsize$/+$5.4}}\\

    \midrule
    \multirow{7}{*}{\textit{4B}}& AIME25 & 23.3& 22.5 & 22.5 & \multirow{5}{*}{OOM}& \textbf{30.4}& \textbf{28.3} \\
    & OlympiadBench & 47.7& 45.0 & 50.1 & & \textbf{50.5}& \textbf{52.0} \\
    & AMC23 & 62.8& 60.9 & 64.1 & & \textbf{70.3}& \textbf{70.6} \\
    & MATH500 & 82.8& 76.1 & 83.2 & & \textbf{84.4}& \textbf{85.6} \\
    & GSM8K & 90.5& 85.3 & \textbf{90.9} & & 90.4& \textbf{91.5} \\
    \cmidrule(lr){2-8}
    & Average & 61.4& 58.0 & 62.2& --& \textbf{65.2{\scriptsize$/+$3.8}}& \textbf{65.6{\scriptsize$/+$4.2}} \\
    \bottomrule
    \end{tabular}
\end{table*}

\xhdr{Math benchmark evaluation}
In this section, all models are trained on the math subset of Open-R1 and evaluated on math benchmarks (see Section~\ref{sec:appendix/exp_setup/training_recipe} for training details).
We also add a \emph{Routing} baseline, which selects a model from a candidate pair for each question. In our experiments, we use two pairs: (1) MobileLLM-R1-360M, Qwen3-1.7B, and (2) Qwen3-0.6B, Qwen3-4B.
All candidate models are SFT-trained under the same settings. For each pair, the routing ratio is calibrated so that the average active parameter count matches our 0.6B and 1.7B targets, respectively.
Table~\ref{tab:math_performance} reports results for 0.6B, 1.7B, and 4B backbones across five challenging math benchmarks.
Even with strong Qwen3-Base initialization, existing approaches show limited effectiveness: AlwaysThink and routing methods fail to consistently outperform the standard baseline, while SoftThink yields only marginal gains.
In contrast, \name achieves stable improvements, with average gains of 4.0\% (0.6B), 5.0\% (1.7B), and 3.8\% (4B) over Standard.
TaH+, which adds less than 3\% more parameters, further improves to 5.3\%, 5.4\%, and 4.2\%, respectively.
For 0.6B and 1.7B, \name achieves 8.1-11.3\% gains over AlwaysThink, and TaH+ achieves 8.5-12.6\% gains. AlwaysThink-4B is not evaluated due to out-of-memory during training.

\xhdr{Out-of-domain evaluation}
We further evaluated the zero-shot generalization capability of models trained solely on math datasets from the main paper. 
As shown in Table~\ref{tab:math_1.7_peformance}, TaH+ demonstrates consistent improvements not only on in-domain math benchmarks (MATH500, AMC23) but also on out-of-domain tasks like MMLU-STEM. This indicates that the thinking patterns learned by TaH+ on math problems are robust and transferrable to broader scientific reasoning tasks.

\begin{table}[tb]
    \centering
    \caption{Performance of math-only trained models (0.6B and 1.7B) on in-domain math benchmarks and the out-of-domain STEM benchmark (MMLU-STEM).}
    \label{tab:math_1.7_peformance}
    \setlength{\tabcolsep}{4pt}
    \begin{tabular}{ll|cccc}
    \toprule
    \textbf{Param.} & \textbf{Benchmark} & \textbf{Standard} & \textbf{SoftThink} & \textbf{AlwaysThink} & \textbf{TaH+} \\
    \midrule
    \multirow{4}{*}{\textit{0.6}} 
        & MATH500    & 47.2 & 48.8 & 32.8 & \textbf{54.2} \\
        & AMC23      & 23.4 & 24.1 & 15.6 & \textbf{30.6} \\
        & MMLU-STEM  & 51.6 & 51.4 & 42.6 & \textbf{56.3} \\
        \cmidrule(lr){2-6}
        & Average    & 40.7 & 41.4 & 30.3 & \textbf{47.0} \\
    \midrule
    \multirow{4}{*}{\textit{1.7}} 
        & MATH500    & 68.4 & 68.8 & 61.8 & \textbf{73.0} \\
        & AMC23      & 42.2 & 43.1 & 42.5 & \textbf{51.2} \\
        & MMLU-STEM  & 70.8 & 70.6 & 63.8 & \textbf{73.7} \\
        \cmidrule(lr){2-6}
        & Average    & 60.5 & 60.8 & 56.0 & \textbf{66.0} \\
    \bottomrule
    \end{tabular}
\end{table}

\subsubsection{Additional Oracle Analysis}
\label{sec:appendix/generation_oracle}

We provide two complementary oracle analyses.
(1) \textbf{Downstream generation.}
Table~\ref{tab:generation_oracle_acc} reports generation results on MATH100 with Qwen3-0.6B.
We use DeepSeek-R1-Distill-Qwen-32B predictions~\citep{guo2025deepseek-r1} as proxy labels.
The oracle verbalizes when the model's top-1 prediction matches the proxy label, and iterates otherwise.
With our trained iteration decider approximating the oracle, \name outperforms both Standard and AlwaysThink baselines.
However, the ideal oracle policy achieves even higher gains, indicating future potential.
(2) \textbf{Next-token prediction.}
Figure~\ref{fig:generation_oracle_transition} shows prediction transitions obtained by verbalizing tokens from all iteration depths.
It reveals that AlwaysThink produces more incorrect than correct revisions, demonstrating latent overthinking.
In contrast, oracle-controlled iterations substantially increase correct revisions by selectively targeting hard tokens.

\begin{center}
    \begin{minipage}[t]{0.42\textwidth}
        \vspace{0pt}
        \captionsetup{hypcap=false}
        \captionof{table}{MATH100 accuracy under different training and inference policies on Qwen3-0.6B, using DeepSeek-R1-Distill-Qwen-32B predictions as proxy labels.}
        \centering
        \setlength\tabcolsep{4pt}
        \begin{tabular}{llc}
        \toprule
        \textbf{Training} & \textbf{Inference} & \textbf{Accuracy} \\
        \midrule
        Standard    & Standard    & 52 \\
        \midrule
        AlwaysThink & AlwaysThink & 38{\scriptsize$/-$14} \\
        AlwaysThink & TaH-Oracle       & 77{\scriptsize$/+$25} \\
        \midrule
        TaH-Oracle   & TaH-Decider   & 54{\scriptsize$/+$\phantom{0}2} \\
        TaH-Oracle   & TaH-Oracle       & 80{\scriptsize$/+$28} \\
        \bottomrule
        \end{tabular}
        \label{tab:generation_oracle_acc}
    \end{minipage}
    \hfill
    \begin{minipage}[t]{0.54\textwidth}
        \vspace{0pt}
        \captionsetup{hypcap=false}
        \captionof{figure}{Next-token prediction transitions across iteration depths on Qwen3-0.6B.}
        \centering
        \includegraphics[width=\linewidth]{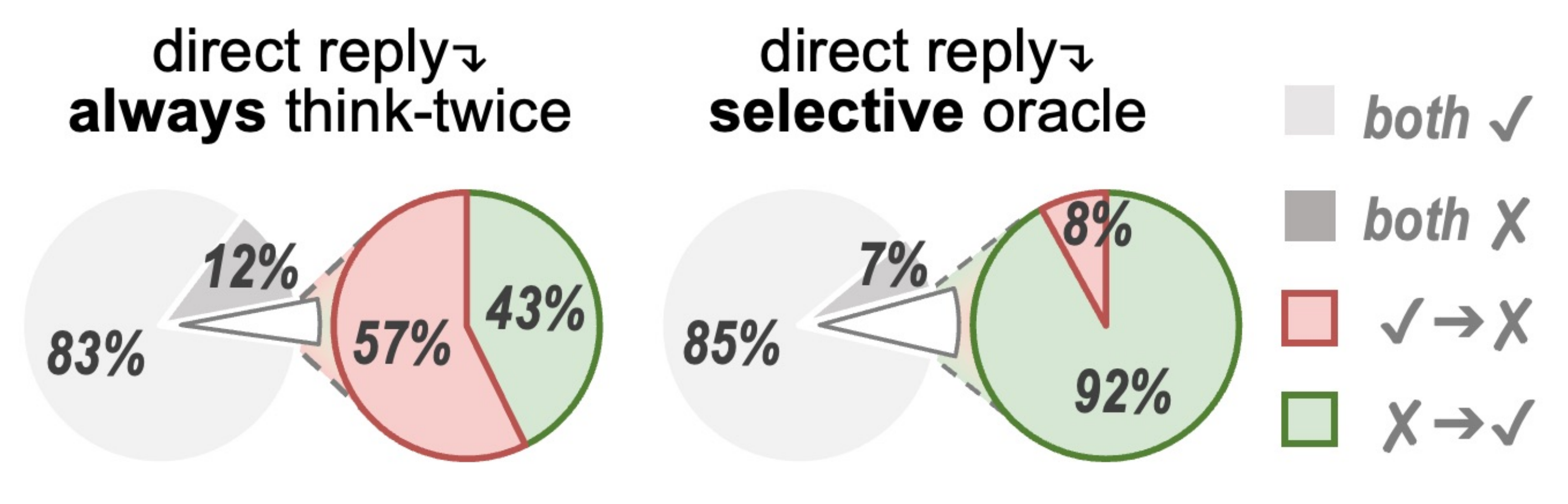}
        \label{fig:generation_oracle_transition}
    \end{minipage}
\end{center}

\subsubsection{Additional Baselines}
\xhdr{Standard-pruned baseline}
Since \name removes one backbone layer to match the parameter budget of Standard, we additionally compare against a Standard-Pruned baseline with the same layer removed but without selective iteration.
Table~\ref{tab:standard_pruned} shows that pruning alone slightly degrades performance, while \name substantially outperforms both Standard and Standard-Pruned under the same training and evaluation setup.
This isolates the gain from selective latent iteration rather than from pruning.

\begin{table}[tb]
    \centering
    \caption{Comparison with a Standard-Pruned baseline on 1.7B math-trained models. Standard-Pruned removes the same backbone layer as \name but does not use latent iteration.}
    \label{tab:standard_pruned}
    \setlength{\tabcolsep}{5pt}
    \begin{tabular}{lcccccc}
    \toprule
    \textbf{Method} & \textbf{AIME25} & \textbf{Olympiad} & \textbf{AMC23} & \textbf{MATH500} & \textbf{GSM8K} & \textbf{Average} \\
    \midrule
    Standard & 13.3 & 33.0 & 42.2 & 68.4 & 82.1 & 47.8 \\
    Standard-Pruned & 11.7 & 32.7 & 41.3 & 68.0 & 79.4 & 46.6 \\
    \name & \textbf{17.9} & \textbf{38.8} & \textbf{48.4} & \textbf{74.4} & \textbf{84.5} & \textbf{52.8} \\
    \bottomrule
    \end{tabular}
\end{table}

\xhdr{Additional latent thinking baselines}
Some latent thinking methods require pre-training and use base models other than Qwen3.
We also compare with these methods, including Ponder~\citep{zeng2025pondering}. Specifically, we adopt the released pretrained PonderingPythia-1.4B as the base model and perform SFT on the same training data. We observe that the fine-tuned model learns the stylistic patterns of the training data,
but still underperforms substantially, which may be attributable to the limited capability of the PonderingPythia-1.4B backbone.

\begin{table}[tb]
\caption{Performance on MATH500 and GSM8K-500 (first 500 GSM8K samples)}
\label{tab:ponder_performance}
\centering
    \begin{tabular}{l|cc}
    \toprule
     & \multicolumn{2}{c}{\textbf{Method}} \\
     \cmidrule(lr){2-3}
     \textbf{Dataset} & Standard-0.6B & Ponder-1.4B \\
    \midrule
     MATH500 & 47.2 & 2.0 \\
     GSM8K-500 & 62.8 &  1.8  \\
    \midrule
     Avg. & 55 &  1.9   \\
    \bottomrule
    \end{tabular}
\end{table}

\subsubsection{Real-world Efficiency}
\label{sec:appendix/exp/real_world_efficiency}

\xhdr{Setup}
We investigate the real-world efficiency of different 1.7B models under our current implementation.
All measurements were obtained on a single A800 GPU with a batch size of 1 and a maximum output length of 8192 tokens, using a challenging AIME25 problem where all three methods reached the token limit.
Memory usage was profiled using \texttt{torch.cuda.memory.\_record\_memory\_history}.

\xhdr{Latency breakdown}
We report the decoding latency, throughput, and a detailed time breakdown for Standard, AlwaysThink, and \name in Table~\ref{tab:latency_breakdown}.
Here, \textit{Iter-1 forward} and \textit{Iter-2 forward} denote the total forward-pass time spent on the first and second latent iterations, respectively; \textit{Iter decider} is the time for the iteration decider network to judge whether to continue iterating or verbalize; \textit{LoRA switching} is the overhead of switching LoRA adapters; and \textit{Other} includes tensor initialization, concatenation, and related bookkeeping.

\begin{table}[t]
    \caption{Per-component latency breakdown on a single A800 GPU.}
    \label{tab:latency_breakdown}
    \centering
    \setlength\tabcolsep{4pt}
    \begin{tabular}{l|cc|cc|cc}
    \toprule
    \textbf{Component}
        & \multicolumn{2}{c|}{\textbf{Standard}}
        & \multicolumn{2}{c|}{\textbf{\name}}
        & \multicolumn{2}{c}{\textbf{AlwaysThink}} \\
        & \textbf{Latency (s)} & \textbf{Ratio(\%)} 
        & \textbf{Latency (s)} & \textbf{Ratio(\%)} 
        & \textbf{Latency (s)} & \textbf{Ratio(\%)} \\
    \midrule
    Iter-1 Forward & 210.6 & 100.0 & 229.8 & 76.2 & 224.1 & 30.0 \\
    Iter-2 Forward &  --   &  --   &  29.6 &  9.8 & 384.7 & 51.5 \\
    Iter. Decider   &  --   &  --   &  10.5 &  3.5 &  --   & --   \\
    LoRA Switching    &  --   &  --   &   7.5 &  2.5 &  91.1 & 12.2 \\
    Other          &  --   &  --   &  24.1 &  8.0 &  47.4 &  6.3 \\
    \bottomrule
    \end{tabular}
\end{table}

\xhdr{Discussion}
We note that our current implementation is not yet optimized at the system level, so there remains room for further efficiency improvements.
For example, the \textit{LoRA Switching} and \textit{Other} overheads (bookkeeping) are relatively high due to the Python-level implementation of dynamic control flow.
These engineering optimizations are largely orthogonal to the algorithmic design of \name, and we plan to continue refining the implementation to further reduce latency and memory overhead.
The theoretical FLOPs and memory access analysis of \name are provided in Appendix~\ref{sec:appendix/exp/theo_eff}.

\subsubsection{Theoretical Efficiency Analysis}
\label{sec:appendix/exp/theo_eff}

Following prior work~\cite{hoffmann2022training, yang2024glitches, wenheng2025cdllm}, we analyze the per-token decoding computation and memory access via analytical operator-level profiling: we trace the forward pass symbolically, accumulating FLOPs and memory bytes from tensor shapes and data types across all modules.

\xhdr{Notation and formulation}
Table~\ref{tab:eff_notation} lists the symbols used throughout this analysis;
Table~\ref{tab:op-cost} breaks down the cost from individual operators (\texttt{nn.linear}, \texttt{nn.sdpa} for \texttt{scaled\_dot\_product\_attention}) through per-layer modules to a full decode step. Since $h=n_h h_h$ for our LLMs, some terms are simplified accordingly.
Lightweight vector ops (RMSNorm, RoPE, SiLU, residual add) are also exhaustively profiled and included in the final numbers in Table~\ref{tab:efficiency_analysis_theo}. Since they generally contribute very little to the total cost, they are omitted from the formulas below for brevity.

\xhdr{Standard decoding}
For a standard LLM generating $t_\text{out}$ output tokens from a prefill of $t_\text{in}$ input tokens, the total FLOPs and memory access are:
\begin{align}
  \textstyle\sum_{i=1}^{t_\text{out}} \F_\text{step}(t_\text{in}{+}i{-}1), \qquad
  \textstyle\sum_{i=1}^{t_\text{out}} \M_\text{step}(t_\text{in}{+}i{-}1).
\end{align}
Per-token averages are obtained by dividing by $t_\text{out}$.

\begin{table}[ht]
\centering
\caption{Notation for theoretical efficiency analysis.}
\label{tab:eff_notation}
\footnotesize
\setlength{\tabcolsep}{4pt}
\begin{tabular}{@{}ll|ll@{}}
\toprule
\textbf{Symbol} & \textbf{Description} & \textbf{Symbol} & \textbf{Description} \\
\midrule
$h$   & Hidden dim              & $n_h$ & Query head count \\
$s$   & KV cache length         & $n_\text{kv}$ & KV head count \\
$d$   & Iteration depth         & $h_h$ & Per-head dim ($h/n_h$) \\
$l$   & Number of layers        & $h_\text{kv}$ & Total KV dim ($n_\text{kv} h_h$) \\
$w$   & Weight count per layer  & $h_\text{ff}$ & FFN intermediate size \\
$b$   & Bytes per element       & $t_\text{in}$ & Input token count \\
$v$   & Vocabulary size         & $t_\text{out}$ & Output token count \\
$k$   & Top-$k$ logits          & & \\
\bottomrule
\end{tabular}
\end{table}

\begin{table}[ht]
\centering
\caption{FLOPs and memory access per decode token, from operators to a full decode step. The Qwen architecture uses GQA ($g{=}n_h/n_\text{kv}$) and gated FFN (SwiGLU). Self-Attention applies 4 \texttt{nn.linear} (Q,K,V,O) plus \texttt{nn.sdpa}; FFN applies 3 \texttt{nn.linear} (gate, up, down). $w{=}2h(h{+}h_\text{kv}){+}3hh_\text{ff}$. Minor terms are absorbed into $\mathcal{O}(\cdot)$.}
\label{tab:op-cost}
\footnotesize
\setlength{\tabcolsep}{4pt}
\begin{tabular}{@{}lll@{}}
\toprule
\textbf{Component} & \textbf{FLOPs} & \textbf{Memory (bytes)} \\
\midrule
\multicolumn{3}{@{}l}{\textit{Operators}} \\[2pt]
\quad \texttt{nn.linear}
  & $2\,h_\text{in}\,h_\text{out}$
  & $(h_\text{in} h_\text{out} {+} h_\text{in} {+} h_\text{out})\,b$ \\[3pt]
\quad \texttt{nn.sdpa}
  & $4\,s\,h$
  & $[2h {+} s(2h_\text{kv} {+} 3n_h)]\,b$ \\
\midrule
\multicolumn{3}{@{}l}{\textit{Modules (per layer)}} \\[2pt]
\quad FFN
  & $6\,h\,h_\text{ff}$
  & $3\,h\,h_\text{ff}\,b + \mathcal{O}(hb)$ \\
\quad Self-Attention
  & $4h(h{+}h_\text{kv}) {+} 4sh$
  & $[2h(h{+}h_\text{kv}) {+} s(2h_\text{kv} {+} 3n_h)]\,b {+} \mathcal{O}(hb)$ \\[3pt]
\midrule
\multicolumn{3}{@{}l}{\textit{Decode step ($\times l$ layers, at length $s$)}} \\[2pt]
\quad Step
  & $l[4h(h{+}h_\text{kv}) {+} 4sh {+} 6hh_\text{ff}] {+} \mathcal{O}(lh_\text{ff})$
  & $l[w {+} s(2h_\text{kv} {+} 3n_h)]\,b {+} \mathcal{O}(lhb)$ \\
\bottomrule
\end{tabular}
\end{table}

\xhdr{Extension to \name}
\name assigns each of the $t_\text{out}$ output tokens ($i=1,\dots,t_\text{out}$) a depth $d_i\in\{1,\dots,d_\text{max}\}$.
Generating one output token at depth $d_i$ requires $d_i$ backbone passes, $d_i{-}1$ inner transitions (Equation~\ref{eq:ours_update}), and $\min(d_i,\,d_\text{max}{-}1)$ decider calls (one after each backbone pass; the call at $d_\text{max}$ is skipped because the token must verbalize).
The total decoding FLOPs and memory access over $t_\text{out}$ output tokens are:
\begin{align}
  &\sum_{i=1}^{t_\text{out}}\Big[\,
    \textstyle\sum_{d=1}^{d_i}\F_\text{step}\big(s_i^{(d)}\big)
    + (d_i{-}1)\,\F_\text{it}+\min(d_i,d_\text{max}{-}1)\,\F_\text{dec}
  \,\Big], \label{eq:tah-flops} \\[4pt]
  &\sum_{i=1}^{t_\text{out}}\Big[\,
    \textstyle\sum_{d=1}^{d_i}\M_\text{step}\big(s_i^{(d)}\big)
    + (d_i{-}1)\,\M_\text{it}+\min(d_i,d_\text{max}{-}1)\,\M_\text{dec}
  \,\Big], \label{eq:tah-mem}
\end{align}
where $s_i^{(d)}$ is the KV cache length visible at depth $d$ for the $i$-th output token under duo-causal attention (Equation~\ref{eq:duo_causal_attention}).
Since the KV cache at each depth is visible to all deeper levels, for $d_\text{max}{=}2$ we have
$s_i^{(1)} = t_\text{in} + i - 1$ (all prior tokens pass through depth~1) and
$s_i^{(2)} = s_i^{(1)} + |\{j < i : d_j {=} 2\}|$ (depth-1 entries plus prior depth-2 tokens).
Table~\ref{tab:tah-extra} lists the per-call cost of the inner transition and iteration decider; both are negligible relative to one backbone pass.
The overhead is therefore dominated by the average depth $\bar{d}=\frac{1}{t_\text{out}}\sum_i d_i$, giving approximately $\bar{d}\times$ the standard cost when $\bar{d}$ is small.

\xhdr{Extension to AlwaysThink} For AlwaysThink, every output token iterates to the maximum depth. When $d_i{=}d_\text{max}{=}2$ for all $i$, every depth-1 KV entry from output tokens has a corresponding depth-2 entry, giving $s_i^{(2)} \approx 2\,s_i^{(1)}$.
The two backbone passes incur $2\times$ the single-pass FFN cost, while their combined attention cost scales as $\mathcal{O}(s_i^{(1)}) + \mathcal{O}(s_i^{(2)}) \approx 3\,\mathcal{O}(s_i^{(1)})$, i.e., ${\sim}3\times$ the single-pass attention FLOPs.
Combining both components yields $2$-$3\times$ total overhead over Standard.

\begin{table}[ht]
\centering
\caption{Per-call cost of \name-specific components. The inner transition is applied $d_i{-}1$ times and the decider $\min(d_i,\,d_\text{max}{-}1)$ times per token. Both are negligible relative to one backbone pass.}
\label{tab:tah-extra}
\footnotesize
\setlength{\tabcolsep}{4pt}
\begin{tabular}{@{}llll@{}}
\toprule
\textbf{Component} & \textbf{Description} & \textbf{FLOPs} & \textbf{Memory (bytes)} \\
\midrule
Inner transition
  & $x_i^{(d+1)}{=}p_i^{(d)}E$ (Eq.~\ref{eq:ours_update})
  & $v + 2hk$
  & $(v {+} kh {+} h)\,b$ \\[3pt]
Iter.\ decider $\mathcal{I}_\phi$
  & MLP ($|\phi|$ params)
  & $2|\phi|$
  & $(|\phi| {+} \text{acts})\,b$ \\
\bottomrule
\end{tabular}
\end{table}

\xhdr{Results}
Tables~\ref{tab:token_iter_stats} and~\ref{tab:token_iter_stats_math} report the average number of input/output tokens and latent iterations per token across benchmarks.
We plug these statistics into the analysis to obtain the per-token costs in Table~\ref{tab:efficiency_analysis_theo}.
With only a small fraction of tokens iterating twice, \name incurs only 1.04-1.05$\times$ the cost of the Standard baseline.
In contrast, AlwaysThink sets $\bar{d}{=}2$, requiring 2.19-2.27$\times$ more computation and memory access.
These results confirm that \name achieves the performance benefits while avoiding the substantial efficiency penalty.

\begin{table}[t]
    \caption{Input tokens (shared across methods) and output token / iteration statistics for Standard, AlwaysThink, \name, and TaH+ (General-trained version).}
    \setlength\tabcolsep{4pt}
    \label{tab:token_iter_stats}
    \centering
    \begin{tabular}{lll|cc|cc|cc|cc|cc}
    \toprule
    & ~ & ~ & \multicolumn{2}{c|}{\textbf{Standard}} & \multicolumn{2}{c|}{\textbf{AlwaysThink}} &  \multicolumn{2}{c|}{\textbf{Ouro}}& \multicolumn{2}{c|}{\textbf{\name}} & \multicolumn{2}{c}{\textbf{TaH+}} \\
    \cmidrule(lr){4-5}\cmidrule(lr){6-7}\cmidrule(lr){8-9}\cmidrule(lr){10-11}\cmidrule(lr){12-13}
    \textbf{Param.} & \textbf{Dataset} & \textbf{In.} & Out. & Iter. & Out. & Iter.  & Out. &Iter.  & Out. & Iter. & Out. & Iter. \\
    \midrule
    \multirow{10}{*}{\textit{0.6B}}& AIME25        & 159& 7687& 1.00& 6885& 2.00& 7370&2.00& 7741& 1.04& 7677& 1.05\\
    & OlympiadBench & 100& 6937& 1.00& 6175& 2.00& 6733&2.00& 6926& 1.04& 6796& 1.04\\
    & AMC23         & 85& 6602& 1.00& 6332& 2.00& 6355&2.00& 6626& 1.04& 6688& 1.04\\
    & MATH500       & 71& 5364& 1.00& 5072& 2.00& 5274&2.00& 5189& 1.04& 5092& 1.04\\
    & GSM8K         & 61& 2144& 1.00& 1710& 2.00& 2039&2.00& 1895& 1.06& 1808& 1.06\\
    & GPQA          & 234& 6533& 1.00& 5998& 2.00& 6087&2.00& 6104& 1.11& 5966& 1.12\\
    & MMLU-STEM     & 97& 2302& 1.00& 2134& 2.00& 2201&2.00& 2057& 1.14& 2033& 1.14\\
    & HumanEval++   & 133& 4711& 1.00& 4411& 2.00& 4149&2.00& 4045& 1.05& 3621& 1.05\\
    & MBPP++        & 51& 4652& 1.00& 4964& 2.00& 4521&2.00& 3094& 1.05& 3175& 1.05\\
    \cmidrule(lr){2-13}
    & Average ratio & -- & 1.00$\times$ &1.00$\times$ & 0.93$\times$ & 2.00$\times$& 0.95$\times$& 2.00$\times$& 0.93$\times$& 1.06$\times$ & 0.91$\times$ &  1.07$\times$\\

    \midrule
    \multirow{10}{*}{\textit{1.7B}}& AIME25        & 159& 7436& 1.00& 7458& 2.00& 7621&2.00& 7799& 1.06& 7547& 1.07\\
    & OlympiadBench & 100& 6190& 1.00& 6318& 2.00& 6505&2.00& 6555& 1.06& 6403& 1.07\\
    & AMC23         & 85& 5918& 1.00& 5766& 2.00& 6261&2.00& 6423& 1.06& 5989& 1.07\\
    & MATH500       & 71& 4159& 1.00& 4347& 2.00& 4455&2.00& 4474& 1.06& 4076& 1.07\\
    & GSM8K         & 61& 1540& 1.00& 1775& 2.00& 1745&2.00& 1718& 1.07& 1597& 1.09\\
    & GPQA          & 234& 6269& 1.00& 6400& 2.00& 6380&2.00& 6729& 1.16& 6502& 1.18\\
    & MMLU-STEM     & 97& 1774& 1.00& 1729& 2.00& 2005&2.00& 1893& 1.14& 1769& 1.18\\
    & HumanEval++   & 133& 3875& 1.00& 3826& 2.00& 4177&2.00& 3905& 1.07& 3797& 1.09\\
    & MBPP++        & 51& 3438& 1.00& 3834& 2.00& 3682&2.00& 3181& 1.07& 3035& 1.09\\
    \cmidrule(lr){2-13}
    & Average ratio & --& 1.00$\times$ & 1.00$\times$& 1.02$\times$& 2.00$\times$& 1.05$\times$ &2.00$\times$& 1.05$\times$ & 1.08$\times$ & 1.00$\times$ & 1.10$\times$ \\

    \midrule
    \multirow{10}{*}{\textit{4B}}& AIME25        & 159& 7007& 1.00& --& --& 7159&2.00& 7452& 1.04& 7282& 1.04\\
    & OlympiadBench & 100& 5699& 1.00& --& --& 5649&2.00& 6010& 1.05& 5898& 1.04\\
    & AMC23         & 85& 5155& 1.00& --& --& 5194&2.00& 5624& 1.04& 5638& 1.04\\
    & MATH500       & 71& 3426& 1.00& --& --& 3434&2.00& 3657& 1.05& 3696& 1.04\\
    & GSM8K         & 61& 1345& 1.00& --& --& 1306&2.00& 1564& 1.06& 1498& 1.06\\
    & GPQA          & 234& 6200& 1.00& --& --& 6133&2.00& 6106& 1.12& 6120& 1.12\\
    & MMLU-STEM     & 97& 1589& 1.00& --& --& 1541&2.00& 1632& 1.11& 1577& 1.11\\
    & HumanEval++   & 133& 3047& 1.00& --& --& 2776&2.00& 3312& 1.05& 3030& 1.09\\
    & MBPP++        & 51& 2673& 1.00& --& --& 2402&2.00& 2633& 1.05& 2584& 1.05\\
    \cmidrule(lr){2-13}
   & Average ratio & --& 1.00$\times$& 1.00$\times$& --& -- & 0.98$\times$&2.00$\times$& 1.05$\times$ &1.06$\times$ & 1.03$\times$ & 1.07$\times$\\

    \bottomrule
    \end{tabular}
\end{table}

\begin{table}[t]
    \caption{Input tokens (shared across methods) and output token / iteration statistics for Standard, AlwaysThink, \name, and TaH+ (Math-trained version).}
    \setlength\tabcolsep{4pt}
    \label{tab:token_iter_stats_math}
    \centering
    \begin{tabular}{lll|cc|cc|cc|cc}
    \toprule
    & ~ & ~ & \multicolumn{2}{c|}{\textbf{Standard}} & \multicolumn{2}{c|}{\textbf{AlwaysThink}} & \multicolumn{2}{c|}{\textbf{\name}} & \multicolumn{2}{c}{\textbf{TaH+}} \\
    \cmidrule(lr){4-5}\cmidrule(lr){6-7}\cmidrule(lr){8-9}\cmidrule(lr){10-11}
    \textbf{Param.} & \textbf{Dataset} & \textbf{In.} & Out. & Iter. & Out. & Iter. & Out. & Iter. & Out. & Iter. \\
    \midrule
    \multirow{6}{*}{\textit{0.6B}}& AIME25        & 159& 7450 & 1.00 & 7316 & 2.00 & 7648 & 1.05 & 7486 & 1.06 \\
    & OlympiadBench & 100& 6599 & 1.00 & 6622 & 2.00 & 6631 & 1.09 & 6513 & 1.06 \\
    & AMC23         & 85& 6377 & 1.00 & 6368 & 2.00 & 6242 & 1.05 & 6145 & 1.05 \\
    & MATH500       & 71& 4823 & 1.00 & 5350 & 2.00 & 4877 & 1.05 & 4793 & 1.06 \\
    & GSM8K         & 61& 1955 & 1.00 & 2844 & 2.00 & 1923 & 1.07 & 1791 & 1.07 \\
    \cmidrule(lr){2-11}
    & Average ratio & --& 1.00$\times$& 1.00$\times$& 1.02$\times$& 2.00$\times$& 1.00$\times$& 1.06$\times$& 0.97$\times$& 1.06$\times$\\

    \midrule
    \multirow{6}{*}{\textit{1.7B}}& AIME25        & 159& 7195 & 1.00 & 7173 & 2.00 & 7496 & 1.06 & 7498 & 1.06\\
    & OlympiadBench & 100& 6008 & 1.00 & 6484 & 2.00 & 6387 & 1.06 & 6258 & 1.06 \\
    & AMC23         & 85& 5681 & 1.00 & 7543 & 2.00 & 6122 & 1.04 & 5852 & 1.06 \\
    & MATH500       & 71& 4004 & 1.00 & 4414 & 2.00 & 4233 & 1.06 & 4286 & 1.06 \\
    & GSM8K         & 61& 1451 & 1.00 & 1644 & 2.00 & 1721 & 1.08 & 1686 & 1.08 \\
    \cmidrule(lr){2-11}
    & Average ratio & --& 1.00$\times$& 1.00$\times$& 1.13$\times$& 2.00$\times$& 1.09$\times$& 1.06$\times$ & 1.07$\times$& 1.06$\times$ \\

    \bottomrule
    \end{tabular}
\end{table}

\begin{table}[t]
    \caption{Decoding computation (GFLOPs) and memory access (GB) per output token for Standard, AlwaysThink, \name, and TaH+.}
    \setlength\tabcolsep{3pt}
    \label{tab:efficiency_analysis_theo}
    \centering
    \begin{tabular}{ll|cc|cc|cc|cc}
    \toprule
    & & \multicolumn{2}{c|}{\textbf{Standard}} & \multicolumn{2}{c|}{\textbf{AlwaysThink}} & \multicolumn{2}{c|}{\textbf{\name}} & \multicolumn{2}{c}{\textbf{TaH+}} \\
    \cmidrule(lr){3-4}\cmidrule(lr){5-6}\cmidrule(lr){7-8}\cmidrule(lr){9-10}
    \textbf{Param.} & \textbf{Dataset} & Comp.& Mem.& Comp.& Mem.& Comp.& Mem.& Comp.& Mem.\\

    \midrule
    \multirow{6}{*}{\textit{0.6B}}& AIME25        & 1.47 & 1.38 & 3.35 & 3.14& 1.52 & 1.43& 1.57 & 1.47 \\
    & OlympiadBench & 1.41 & 1.32 & 3.21 & 3.02& 1.51 & 1.42& 1.50 & 1.41 \\
    & AMC23 & 1.40 & 1.31 & 3.17 & 2.97& 1.43 & 1.34& 1.46 & 1.37 \\
    & MATH500 & 1.31 & 1.22 & 2.98 & 2.80& 1.35 & 1.26& 1.39 & 1.31 \\
    & GSM8K & 1.14 & 1.06 & 2.54 & 2.38& 1.19 & 1.12& 1.22 & 1.14 \\
    \cmidrule(lr){2-10}
    & Average ratio & 1.00$\times$ & 1.00$\times$ & 2.27$\times$ & 2.27$\times$& 1.04$\times$ & 1.04$\times$& 1.06$\times$ & 1.06$\times$ \\
    \midrule
    \multirow{6}{*}{\textit{1.7B}}& AIME25        & 4.31 & 4.03 & 9.45 & 8.83& 4.51 & 4.21& 4.64 & 4.34 \\
    & OlympiadBench & 4.16 & 3.88 & 9.18 & 8.58& 4.36 & 4.07& 4.48 & 4.18 \\
    & AMC23 & 4.12 & 3.85 & 9.54 & 8.91& 4.24 & 3.96& 4.43 & 4.13 \\
    & MATH500 & 3.92 & 3.66 & 8.45 & 7.89& 4.10 & 3.83& 4.23 & 3.95 \\
    & GSM8K & 3.62 & 3.38 & 7.48 & 6.98& 3.87 & 3.61& 3.98 & 3.72 \\
    \cmidrule(lr){2-10}
    & Average ratio & 1.00$\times$ & 1.00$\times$ & 2.19$\times$ & 2.19$\times$& 1.05$\times$ & 1.05$\times$& 1.08$\times$ & 1.08$\times$ \\
    \bottomrule
    \end{tabular}
\end{table}





\subsection{Additional Analysis}
\label{sec:appendix/exp/analysis}

\subsubsection{Training Scheme Dynamics}
\label{sec:appendix/exp/training_dynamics}
\begin{figure}[tb]
    \centering
    \begin{subfigure}[t]{0.48\textwidth}
        \centering
        \includegraphics[width=\textwidth]{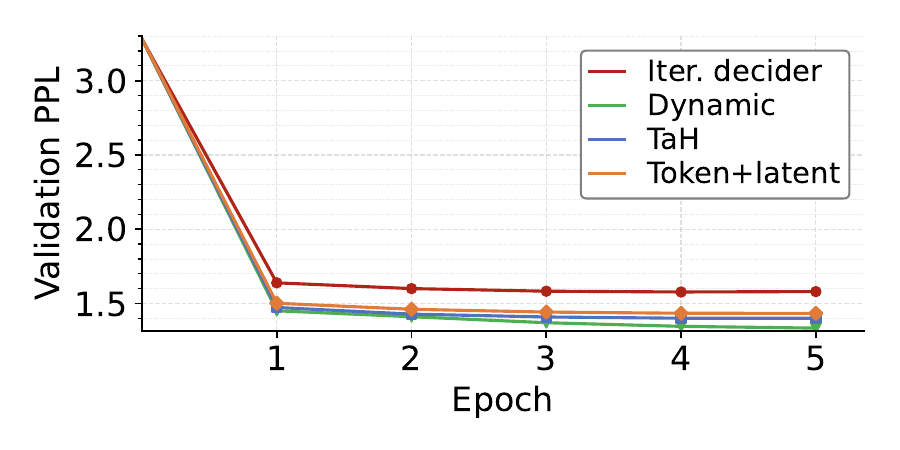}
        \caption{Supervision type and iteration policy.}
        \label{fig:ablation_validation_ppl}
    \end{subfigure}
    \hfill
    \begin{subfigure}[t]{0.48\textwidth}
        \centering
        \includegraphics[width=\textwidth]{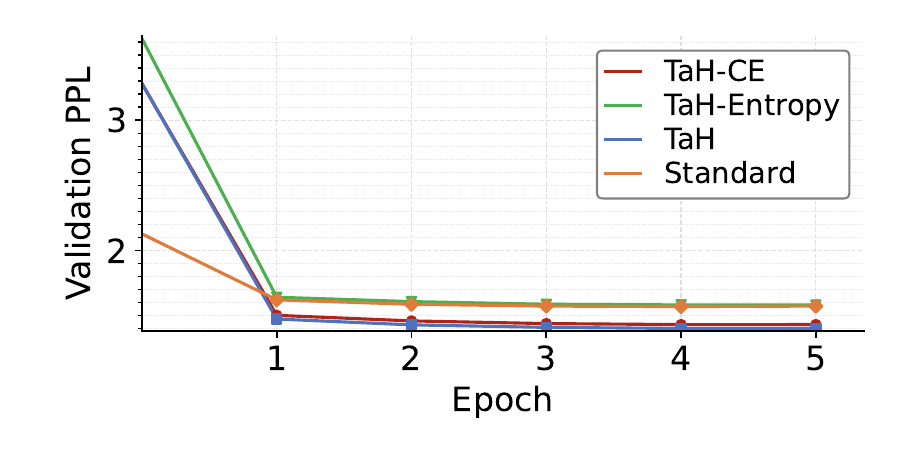}
        \caption{Oracle discrepancy metric.}
        \label{fig:loss_difficulty_metric}
    \end{subfigure}
    \caption{Validation-perplexity dynamics for the training-scheme ablations in Table~\ref{tab:ablation}. (a) Supervision and iteration-policy choices. (b) Oracle discrepancy metrics under matched continuation budgets.}
    \label{fig:training_scheme_dynamics}
\end{figure}

Figure~\ref{fig:training_scheme_dynamics} complements the training-scheme ablations in Table~\ref{tab:ablation} with validation-perplexity curves.
Token-only supervision with a fixed oracle policy trains stably, while token+latent supervision and decider-based training lag behind.
The dynamic oracle can reduce validation perplexity but is unstable during generation, matching its poor downstream accuracy.
For the oracle discrepancy metric, top-1 mismatch gives the best trajectory under matched continuation budgets, supporting our default policy. Note that these training-scheme dynamics reflect a performance ceiling rather than the actual downstream task performance.

\subsubsection{Iteration Label Analysis}
\label{sec:appendix/exp/oracle_label}

\begin{figure}[tb]
    \centering
    \begin{subfigure}[t]{0.32\textwidth}
        \centering
        \includegraphics[width=\textwidth]{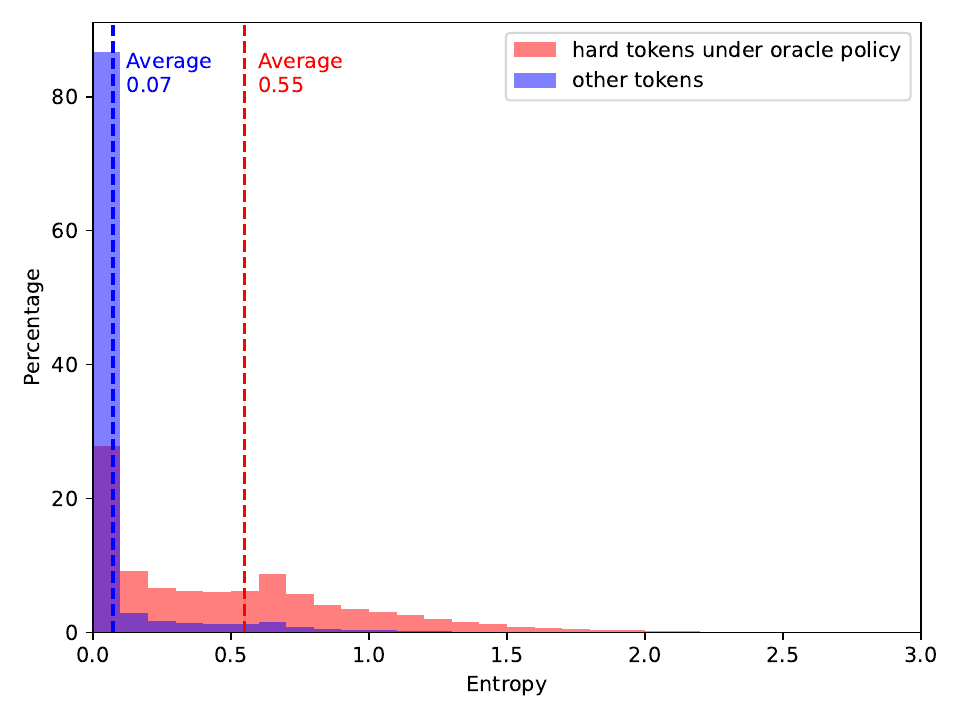}
        \caption{Math}
        \label{fig:label_math}
    \end{subfigure}
    \hfill
    \begin{subfigure}[t]{0.32\textwidth}
        \centering
        \includegraphics[width=\textwidth]{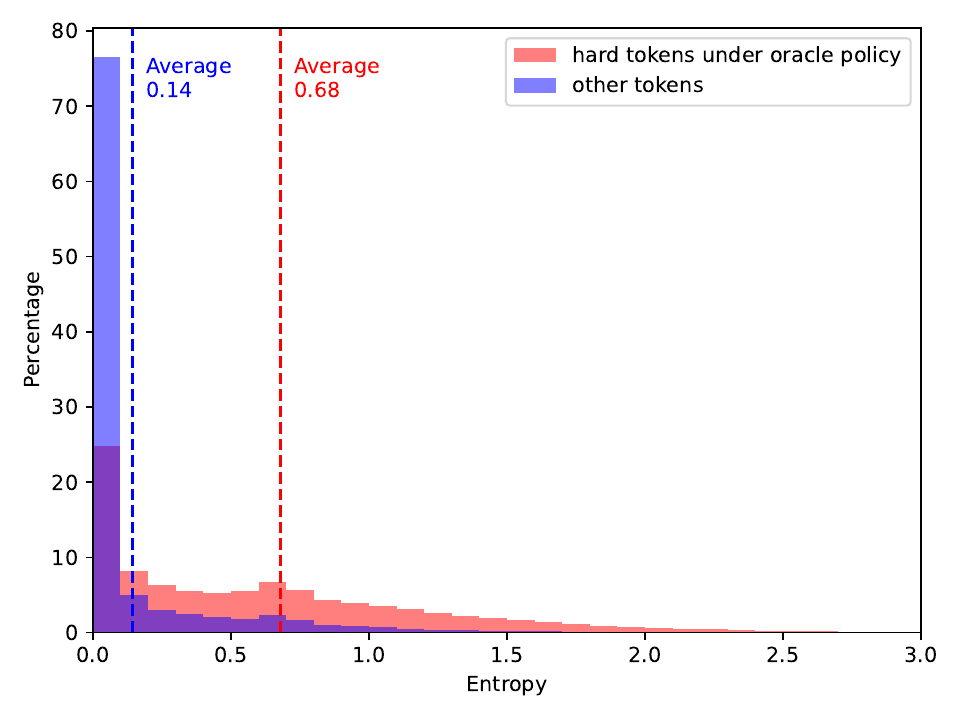}
        \caption{Science}
        \label{fig:label_science}
    \end{subfigure}
    \hfill
    \begin{subfigure}[t]{0.32\textwidth}
        \centering
        \includegraphics[width=\textwidth]{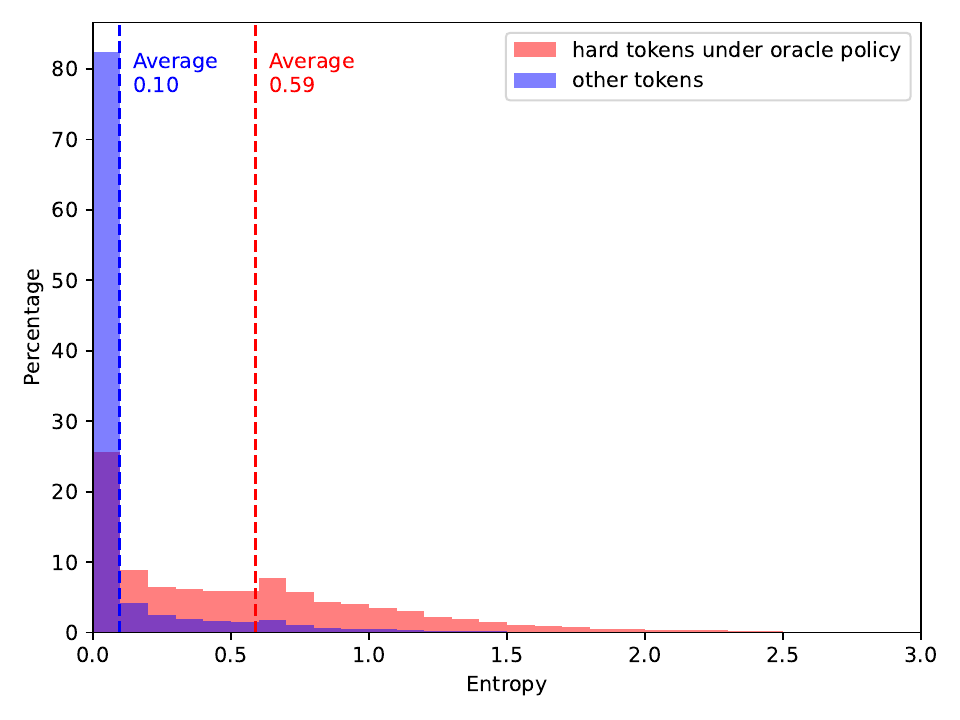}
        \caption{Code}
        \label{fig:label_code}
    \end{subfigure}
    \caption{Output-logit entropy distribution at the first iteration of \name, categorized by oracle iteration labels on the Open-R1 validation set (Math, Science, Code). The distinct separation between distributions confirms that \name's internal logits provide a strong, task-agnostic signal for identifying tokens that require additional iteration.}
    \label{fig:label_hard_tokens}
\end{figure}

\xhdr{Iteration label identifiability}
We first examine whether the oracle iteration labels expose stable token-level signals.
We compute the token entropy of continue-labeled and stop-labeled tokens across three diverse subsets of the Open-R1 dataset (Math, Science, and Code).
As shown in Figure~\ref{fig:label_hard_tokens}, continue-labeled tokens exhibit a universal signature of significantly higher entropy ($>5\times$) than stop-labeled tokens.
This distinct separation confirms that iteration need is an intrinsic, robustly identifiable property of the model's predictive state, rather than a complex, task-specific pattern.
Given this clear signal, the neural iteration decider can learn reliable classification strategies that generalize well across different domains.

\begin{figure}[tb]
    \centering
    \hfill
    \begin{subfigure}[t]{0.48\textwidth}
        \centering
        \includegraphics[width=0.5\textwidth]{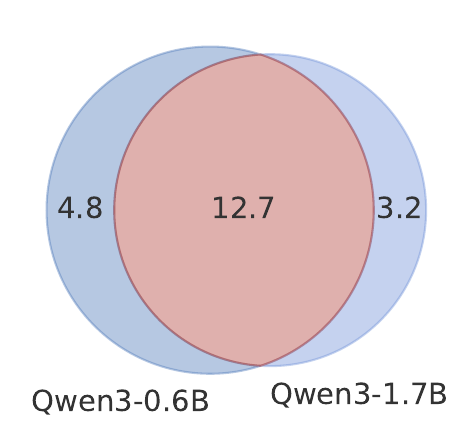}
        \caption{Overlap between Qwen3-0.6B and 1.7B}
        \label{fig:venn_0_6_1_7}
    \end{subfigure}
    \hfill
    \begin{subfigure}[t]{0.48\textwidth}
        \centering
        \includegraphics[width=0.5\textwidth]{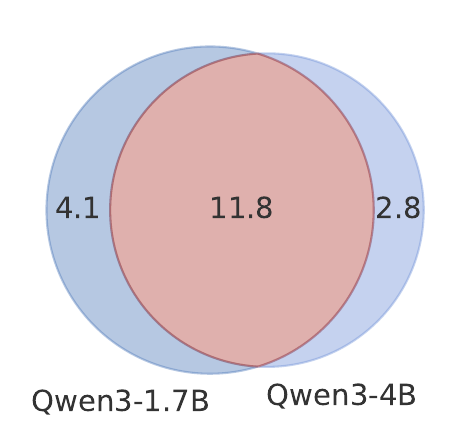}
        \caption{Overlap between Qwen3-1.7B and 4B}
        \label{fig:venn_1_7_4}
    \end{subfigure}
    \hfill
    \caption{Venn diagrams illustrating the overlap of iteration-selected tokens between different reference models. The high overlap proportions indicate that iteration labels are largely consistent across model scales.}
    \label{fig:venn_overlap}
\end{figure}

\begin{figure}[tb]
    \centering
    \begin{subfigure}[t]{0.48\textwidth}
        \centering
        \includegraphics[width=0.7\textwidth]{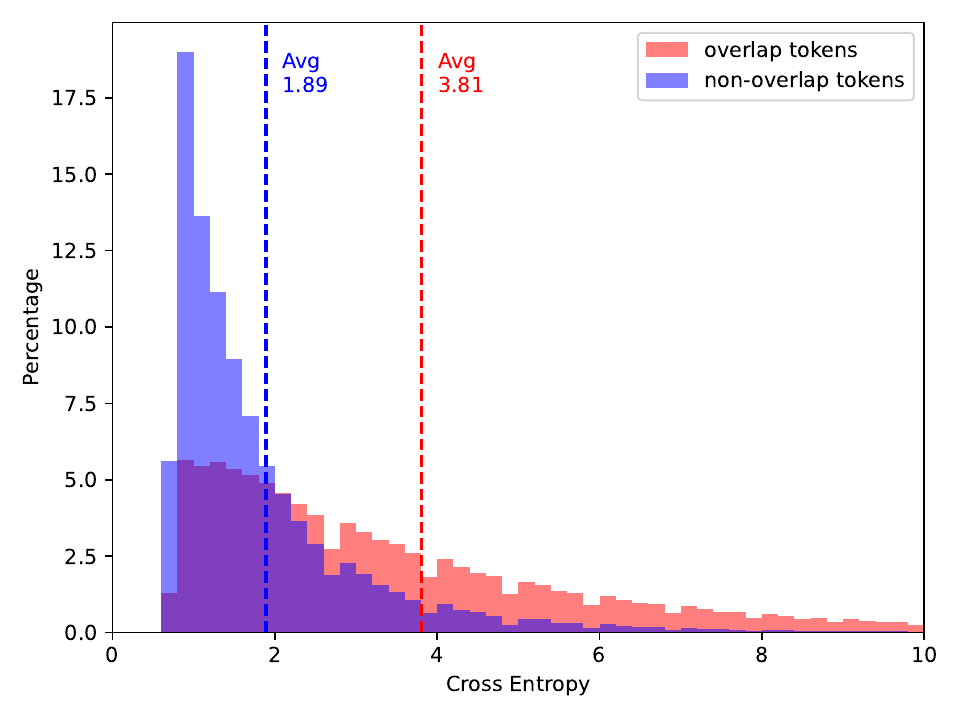}
        \caption{Qwen3-1.7B: cross-entropy of overlap vs. non-overlap iteration-selected tokens (w.r.t. Qwen3-0.6B).}
        \label{fig:ce_1_7_vs_0_6}
    \end{subfigure}
    \hfill
    \begin{subfigure}[t]{0.48\textwidth}
        \centering
        \includegraphics[width=0.7\textwidth]{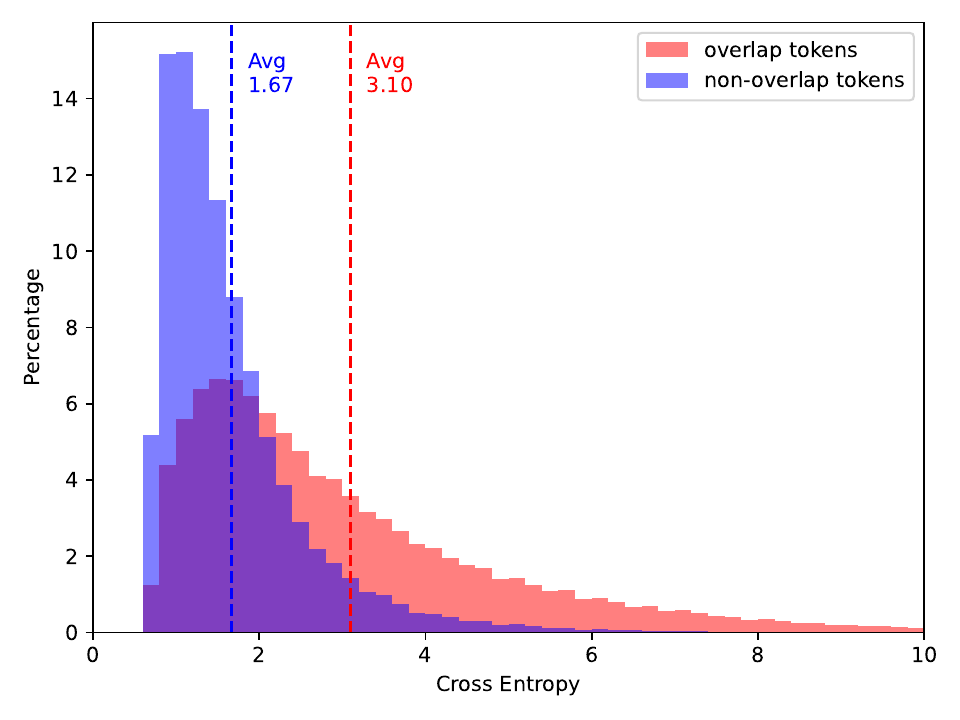}
        \caption{Qwen3-0.6B: cross-entropy of overlap vs. non-overlap iteration-selected tokens (w.r.t. Qwen3-1.7B).}
        \label{fig:ce_0_6_vs_1_7}
    \end{subfigure}

    \vspace{0.4em}

    \begin{subfigure}[t]{0.48\textwidth}
        \centering
        \includegraphics[width=0.7\textwidth]{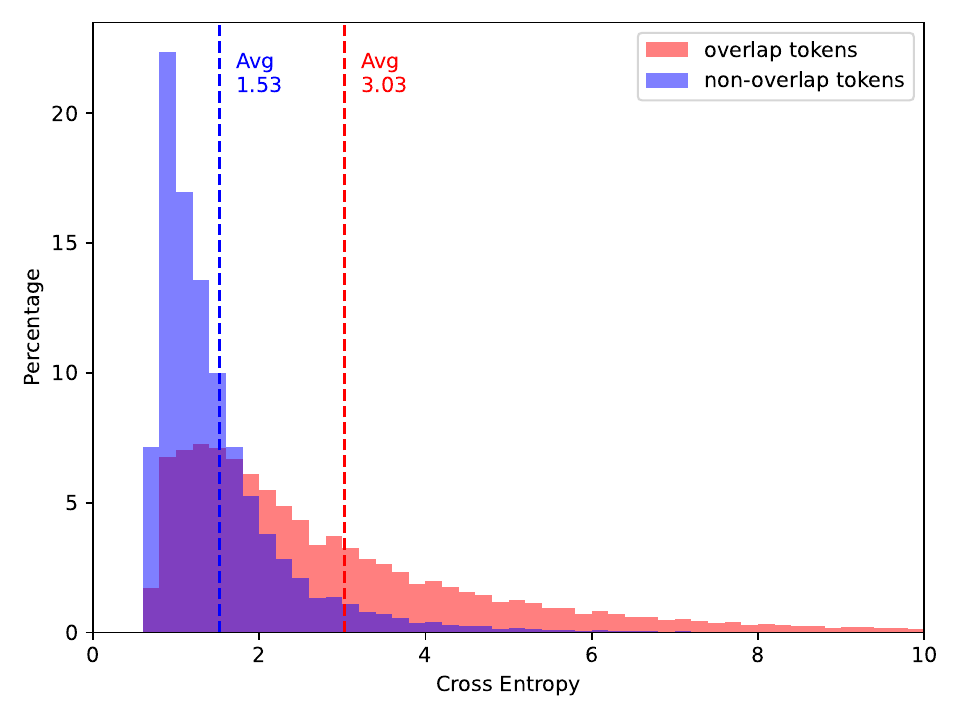}
        \caption{Qwen3-4B: cross-entropy of overlap vs. non-overlap iteration-selected tokens (w.r.t. Qwen3-1.7B).}
        \label{fig:ce_4_vs_1_7}
    \end{subfigure}
    \hfill
    \begin{subfigure}[t]{0.48\textwidth}
        \centering
        \includegraphics[width=0.7\textwidth]{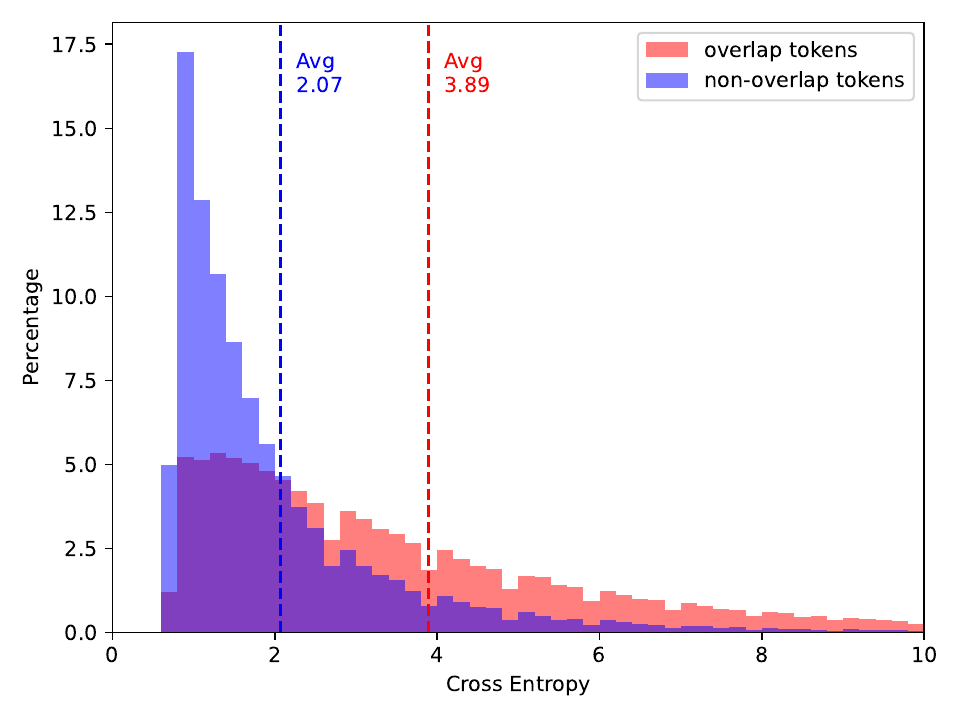}
        \caption{Qwen3-1.7B: cross-entropy of overlap vs. non-overlap iteration-selected tokens (w.r.t. Qwen3-4B).}
        \label{fig:ce_1_7_vs_4}
    \end{subfigure}

    \caption{Token-level cross-entropy distributions of overlap and non-overlap iteration-selected tokens across different reference model pairs. For each pair of reference models (e.g., Qwen3-1.7B and Qwen3-0.6B), we plot the cross-entropy of tokens selected for iteration by both models (overlap) and by only one model (non-overlap) on both reference models.}
    \label{fig:ref_overlap_ce}
\end{figure}

\xhdr{Cross-model label consistency}
We next analyze whether iteration labels are stable across reference model scales.
Figure~\ref{fig:venn_overlap} shows substantial overlap; for example, the 1.7B reference identifies 81\% of the iteration-selected tokens found by the 4B reference.
To assess the quality of this overlap, we compare the cross-entropy of overlap and non-overlap iteration-selected tokens in Figure~\ref{fig:ref_overlap_ce}.
Overlap tokens have consistently higher loss ($\approx 2.0\times$), indicating a stable core of genuinely difficult token positions.
These results show that iteration labels expose stable uncertainty signals, motivating the generalization analysis of the learned iteration decider below.

\subsubsection{Iteration Decider Analysis}

\xhdr{Cross-domain behavior}
We evaluate the iteration decider, trained on the general Open-R1 corpus, across three validation subsets (Math, Code, and QA) to quantify its robustness and cross-domain generalizability.
As summarized in Table~\ref{tab:decider_generalization}, the decider maintains high decision accuracy across all domains without any retraining.

Despite being invoked on only 7.8-26.6\% of tokens, the decider consistently yields 5.2-9.0\% absolute accuracy gains over the standard single-pass baseline on all three domains.
Moreover, the decider automatically adjusts its iteration rate according to task difficulty: it iterates more frequently on QA (26.6\%) than on Math (7.8\%), even under a fixed threshold $c_{\text{threshold}}=0.9$.
This behavior indicates that the decider responds to intrinsic uncertainty signals in the model's predictive distribution rather than memorizing domain-specific patterns.

\xhdr{Iteration decision error}
We further analyze how specific decision mistakes affect end-to-end response quality.
Because the learned decider is imperfect, as shown in Figure~\ref{fig:training_decider}, we randomly inject errors into the oracle iteration-decider predictions at different rates.
Formally, we denote the original oracle prediction as the \textit{label} $l \in \{0,1\}$ and the altered prediction as the \textit{output} $o \in \{0,1\}$.
We define the \textit{iter. error} as the total proportion of deliberately introduced errors:
\begin{equation}
    \text{iter. error} = P(l \neq o) =
    \underbrace{P(l=1,o=0)}_{\textit{underthink rate}}
    + \underbrace{P(l=0,o=1)}_{\textit{overthink rate}}.
\end{equation}
We further distinguish the impacts of overthinking and underthinking.
Here, overthinking refers to cases where the decider incorrectly signals \textit{continue}, while underthinking corresponds to cases where it incorrectly signals \textit{stop}.
Table~\ref{tab:oracle_with_noise} shows how \name's MATH100 accuracy varies with different iteration error rates.
We quantify these effects by fitting a linear model to the data:
$$
\text{accuracy} = -1.41 \times \text{underthink rate} - 2.73 \times \text{overthink rate} + 0.81.
$$
This analysis indicates that inaccurate iteration decisions are the main factor behind the performance gap between \name and its oracle variant, with overthinking being the dominant source of performance gaps.

\begin{table}[tb]
    \centering
    \begin{minipage}[t]{0.47\textwidth}
        \vspace{0pt}
        \centering
        \caption{Iteration decider behavior and downstream gains on different validation subsets. The decider is trained once on general Open-R1 and evaluated without retraining.}
        \label{tab:decider_generalization}
        \footnotesize
        \setlength{\tabcolsep}{3pt}
        \begin{tabular}{lccc}
        \toprule
        \textbf{Metric} & \textbf{Math} & \textbf{Code} & \textbf{QA} \\
        \midrule
        Iteration Percentage & 7.8\% & 10.7\% & 26.6\% \\
        Iteration Accuracy & 86.7\% & 82.3\% & 76.6\% \\
        \makecell[l]{Benchmark Gain\\over Standard} & \textbf{+5.2\%} & \textbf{+9.0\%} & \textbf{+5.8\%} \\
        \bottomrule
        \end{tabular}
    \end{minipage}
    \hfill
    \begin{minipage}[t]{0.50\textwidth}
        \vspace{0pt}
        \centering
        \caption{\name performance under different iteration-decider error rates. All values are reported in percentages.}
        \label{tab:oracle_with_noise}
        \footnotesize
        \setlength{\tabcolsep}{3pt}
        \begin{tabular}{ccc|c}
        \toprule
        \makecell{\textbf{Iter.}\\\textbf{Error (\%)}} & \makecell{\textbf{Underthink}\\\textbf{(\%)}} & \makecell{\textbf{Overthink}\\\textbf{(\%)}} & \makecell{\textbf{MATH100}\\\textbf{Acc. (\%)}}\\
        \midrule
        0.0 & 0.0 & 0.0 & 80.0\\
        2.8 & 2.8 & 0.0 & 78.0\\
        10.0 & 1.5 & 8.5 & 55.4\\
        15.0 & 2.1 & 12.9 & 45.2\\
        20.0 & 2.5 & 17.5 & 27.1\\
        22.1 & 0.0 & 22.1 & 21.6\\
        \bottomrule
        \end{tabular}
    \end{minipage}
\end{table}

\subsubsection{Token Alternation Pattern}
\label{sec:appendix/exp/flow}

We analyze tokens that most frequently trigger a second iteration ("think-twice" tokens). For each token type $t$, we compute the continuation rate
\[
\Pr\big(c_i^{(1)} > c_{\text{threshold}} \mid t_i = t\big),
\]
using the inference threshold $c_{\text{threshold}}=0.9$ (Section~\ref{sec:training}). We estimate this quantity on the Open-R1 validation set and, for diagnostics, randomly sample 10K token positions (\(\approx\)0.4\% of tokens) to track whether the next-token prediction switches between depth 1 and depth 2. This setting quantifies which token types most often trigger an additional iteration and how often iteration alters the predicted next token.

\begin{table}[tb]
    \centering
    \caption{Conditional probabilities of continuation confidence and next-token distribution.}
    \begin{tabular}{lclc}
    \toprule
    \textbf{Token $T_1$} & $P(c^{(1)} > c_{\text{threshold}} \mid t^{(1)}=T_1)$ & \textbf{Token $T_2$} & $P(t^{(2)}=T_2 \mid t^{(1)}=T_1)$ \\
    \midrule
    \multirow{3}{*}{But} & \multirow{3}{*}{34.3\%} & So        & 13.63\% \\
                         &                          & Wait      & 12.17\% \\
                         &                          & Therefore & 8.95\%  \\
    \midrule
    \multirow{3}{*}{So}  & \multirow{3}{*}{17.7\%} & So        & 28.17\% \\
                         &                          & Therefore & 13.67\% \\
                         &                          & But       & 4.89\%  \\
    \bottomrule
    \end{tabular}
    \label{tab:token_probs}
\end{table}

\subsubsection{Duo-causal Attention Pattern}
\label{sec:appendix/exp/attention}

We perform forward computation on 100 samples, each with a length of 128 tokens, and visualize the learned attention patterns of the \name model during the second iteration.

\begin{figure}[tb]
    \centering
    \includegraphics[width=\textwidth]{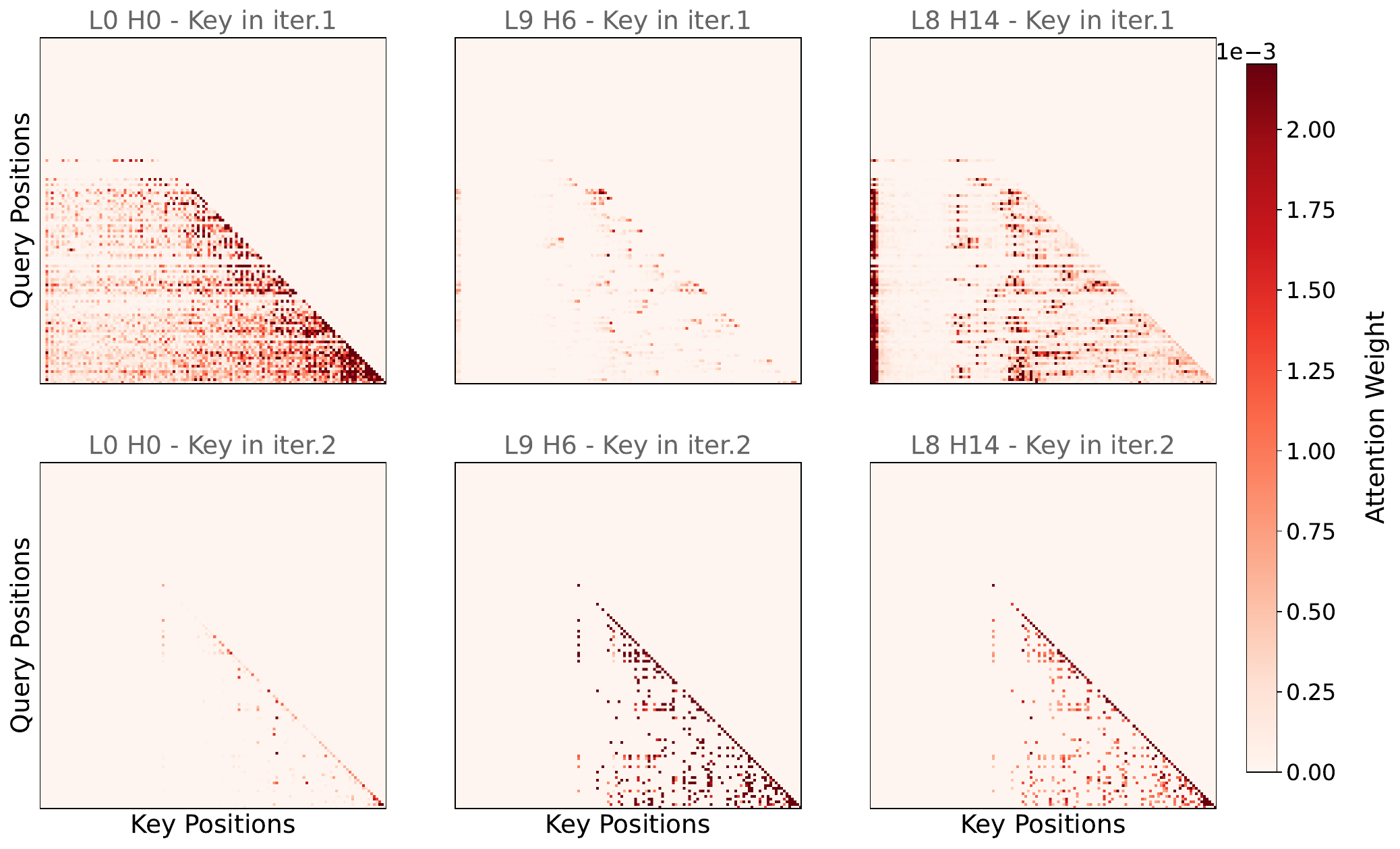}
    \caption{\name duo-causal attention pattern.}
    \label{fig:attention}
\end{figure}

\xhdr{Qualitative analysis}
Figure~\ref{fig:attention} shows the average attention weights of three representative heads.
The left panel illustrates a head that mainly attends to first-iteration keys; the middle panel shows one focusing on second-iteration keys; and the right panel displays a head with balanced attention across both iterations.
These examples demonstrate that the duo-causal attention mechanism enables the model to automatically learn diverse cross-depth attention strategies.

\begin{figure}[tb]
    \centering
    \includegraphics[width=0.55\textwidth]{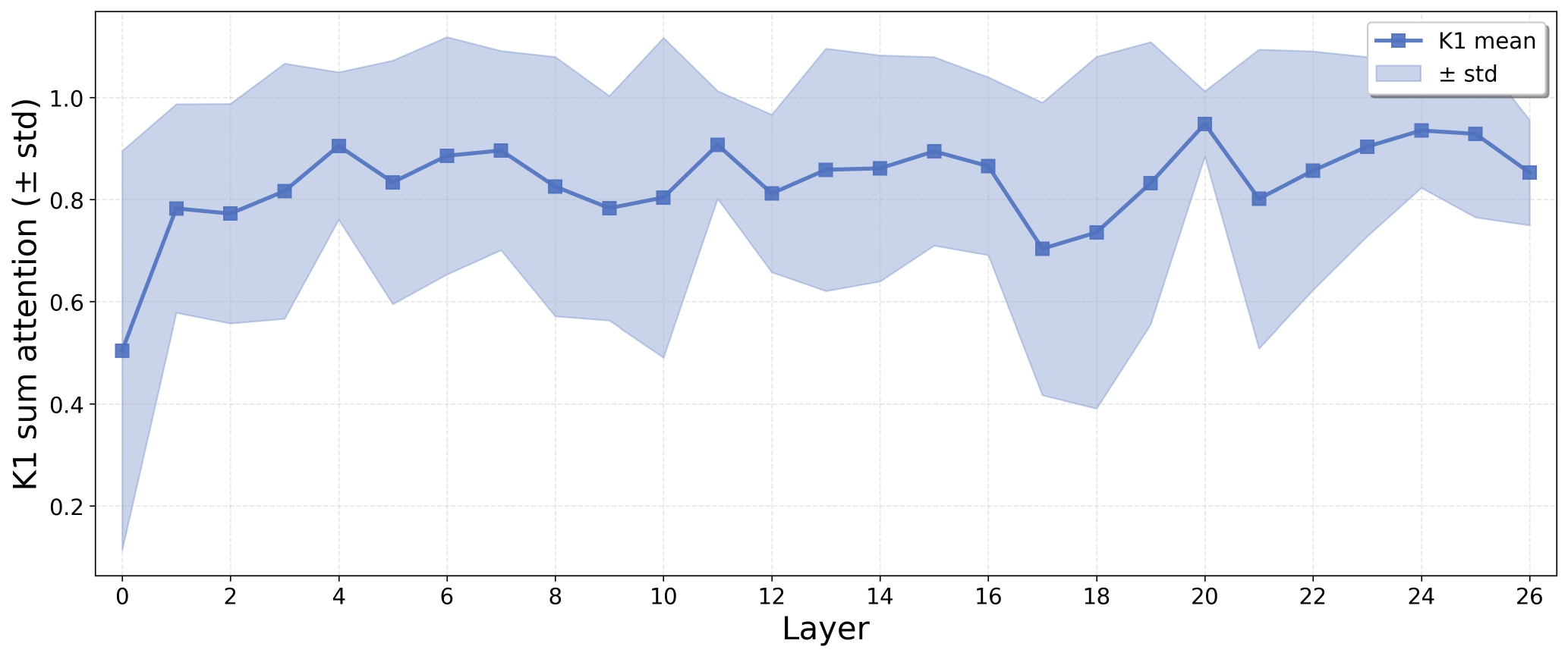}
    \caption{\name layer-wise cross-iteration attention mass $\bar{m}^{(2\to 1)}$ (mean $\pm$ one standard deviation over sequences and heads).}
    \label{fig:K1_attention_layer_curve}
\end{figure}

\xhdr{Quantitative analysis}
Figure~\ref{fig:K1_attention_layer_curve} further quantifies, for each layer, how much attention mass second-iteration queries allocate to first-iteration keys.
Following the \name framework, let $a_{\ell,h}\!\big((i,d_q)\!\to\!(j,d_{kv})\big)$ denote the attention weight at layer $\ell$, head $h$, from the query at position $i$, depth $d_q$ to the key at position $j$, depth $d_{kv}$.
As an illustrative example, consider query $(3,2)$ highlighted in Figure~\ref{fig:duo_attention_expand}(c): it attends to three depth-1 keys and two depth-2 keys. We want to quantify how attention mass is distributed between these two groups.
We define the \emph{cross-iteration attention mass} for a single query position as
\[
m_{i,\ell,h}^{(2\to 1)} \;=\; \sum_{j \le i} a_{\ell,h}\!\big((i,2)\!\to\!(j,1)\big),
\]
and average over all depth-2 query positions within each sequence $s$:
\[
\bar{m}_{s,\ell,h}^{(2\to 1)}
\;=\;
\frac{1}{\lvert Q_s^{(2)} \rvert}
\sum_{i \in Q_s^{(2)}} m_{i,\ell,h}^{(2\to 1)},
\]
where $Q_s^{(2)}$ is the set of token positions that iterate to depth $2$ in sequence $s$.
The curve in Figure~\ref{fig:K1_attention_layer_curve} plots, for each layer, the mean of $\bar{m}_{s,\ell,h}^{(2\to 1)}$ over all sequence-head pairs $(s,h)$, with shading indicating one standard deviation.

\xhdr{Findings}
A higher $\bar{m}^{(2\to 1)}$ indicates that depth-2 queries rely more on first-iteration keys, whereas a lower value indicates greater reliance on same-iteration keys.
The results reveal substantial layer-wise heterogeneity: lower layers place relatively less mass on first-iteration keys, while deeper layers show higher and varied cross-depth reuse.

\subsection{Implementation Details}

\subsubsection{Duo-causal attention implementation}
\label{sec:appendix/duo-causal-attention}

Figure~\ref{fig:duo_attention_expand} illustrates the implementation of duo-causal attention, with the formal definitions provided below.

\textbf{(1) KV cache concatenation.} At depth $d$, we form the visible K/V sequence by concatenating all shallower-to-current depths along the sequence dimension:
\[ \mathrm{KV}^{(\le d)} \,=\, [\, \mathrm{KV}^{(1)}\,;\,\mathrm{KV}^{(2)}\,;\,\cdots\,;\,\mathrm{KV}^{(d)} \,]. \]
This realizes the accessible set in Equation~\ref{eq:duo_causal_attention}, allowing deeper iterations to access all shallower iterations while preserving positional causality. 
The KV cache is managed by iteration depth during decoding, as shown in Figure~\ref{fig:duo_attention_expand}(b). The fragmented KV-cache management strategy is standard in existing LLM serving systems~\citep{kwon2023vllm, zheng2024sglang}.

\textbf{(2) Two-dimensional causal mask.} For a query $(i,d)$, a key $(j,k)$ is attendable iff $j\le i$ and $k\le d$. We implement this as an additive attention mask with $0$ for allowed entries and $-\infty$ otherwise, enforcing positional and iteration causality jointly. 
Figure~\ref{fig:duo_attention_expand}(c) visualizes the landscape of the duo-causal attention mask.
When $d=1$ for all tokens, the rule reduces to standard causal attention.

\textbf{(3) Compatibility with efficient attention.} The mask is provided in the standard additive form and the concatenated K/V tensors remain contiguous along the sequence dimension, matching the usual scaled dot-product attention interface.
As a result, duo-causal attention is directly compatible with optimized kernels such as FlashAttention, without kernel modifications.

\subsection{Additional Related Work}
Instead of using the shared model parameters multiple times through latent iteration, previous work also proposes layer skipping methods for dynamic compute allocation.

\xhdr{Layer skipping}
Layer skipping aims to accelerate LLM inference by dynamically bypassing certain layers for specific tokens.
Some methods use a learnable module to make real-time skipping decisions. MoD~\citep{raposo2024mixture} uses a top-k router to select a subset of tokens for processing, while FlexiDepth~\citep{luo2025adaptive} uses a plug-in router to determine whether a layer should be bypassed. Others use a fixed strategy to skip layers. SkipDecode~\citep{del2023skipdecode} enforces a monotonically decreasing number of active layers during generation.
However, these methods still require loading the entire model's parameters, resulting in a large memory access overhead.
Instead of skipping some layers, \name adds computational depth by allowing core tokens to undergo multiple refinement iterations.
This approach provides greater computational depth without increasing the model's parameter count.


\subsection{Limitations and Future Work}
\xhdr{Comparison with official Qwen3 models}
Official Qwen3 models are trained on different data distributions and scales, and use different training procedures, including on-policy distillation~\citep{yang2025qwen3}. By contrast, our models use SFT only on limited, publicly accessible data. Consequently, performance may differ between the two.

\xhdr{Future work}
We consider advanced training techniques largely orthogonal to \name. Promising directions include (1) integrating \name with online distillation to provide stronger supervision; (2) applying reinforcement learning to optimize the compute–accuracy trade-off and to learn continuation policies beyond imitation. Due to resource constraints, this work focuses on SFT with public data; we leave these combinations for future work.

\subsection{Use of Large Language Models (LLMs)}
Large language models (e.g., GPT-5, Gemini) were used exclusively for proofreading and minor copyediting. They were not employed to generate scientific content, contribute to research ideation, or design methods. All research ideas, methods, and conclusions are solely the responsibility of the authors.

\subsection{Examples}

We use an example question from AIME25 and responses from Standard, TaH and AlwaysThink to provide an intuitive understanding of our method. Only TaH can successfully answer this question. Blue-highlighted tokens indicate positions that undergo two iterations.

\begin{bluebox}[Question]{width=\linewidth, valign=top}
In $\triangle ABC$, points $A,D,E$, and $B$ lie in that order on side $\overline{AB}$ with $AD=4, DE=16$, and $EB=8$. Points $A,F,G$, and $C$ lie in that order on side $\overline{AC}$ with $AF=13, FG=52$, and $GC=26$. Let $M$ be the reflection of $D$ through $F$, and let $N$ be the reflection of $G$ through $E$. Quadrilateral $DEGF$ has area 288. Find the area of the heptagon $AFNBCEM$.
\end{bluebox}

\clearpage

\begin{figure*}[t]
\centering
\begin{minipage}[t]{0.32\textwidth}
\begin{bluebox}[Standard Output]{width=\linewidth, height=0.95\textheight, valign=top}

Okay, let me try to figure out this problem step by step. It's about a heptagon formed by reflecting points and finding its area. \ldots
We have triangle $ABC$. On side $AB$, points $A, D, E$, and $B$ are in that order. The lengths are given as $AD = 4$, $DE = 16$, and $EB = 8$. So $AB = 28$ \ldots
On side $AC$, points $A, F, G$, and $C$ are in that order with $AF = 13$, $FG = 52$, and $GC = 26$. So $AC = 91$ \ldots
Let me place point $A$ at $(0,0)$ and $B$ at $(28,0)$ \ldots Let $C$ be at $(91, h)$ \ldots Then
$D=(4,0)$, $E=(20,0)$, $F=\big(13,\frac{13h}{91}\big)$, $G=\big(65,\frac{65h}{91}\big)$ \ldots

Area of quadrilateral $DEGF$ can be calculated \ldots Applying the shoelace formula \ldots
Area $= \frac{1}{2}\cdot \frac{1248h}{91} = \frac{624h}{91} \;\mathbf{=}\; 6h$

So with $288$:
$6h = 288 \Rightarrow h = \frac{288}{6} = 48$

Reflections: $M$ is the reflection of $D$ through $F$ and $N$ is the reflection of $G$ through $E$ \ldots
$M=(22,0)$, $N=(-39,0)$, $B=(28,0)$, $C=(91,48)$, $E=(20,0)$ \ldots

Using the shoelace formula for $AFNBCEM$ in order \ldots
Sum $= 0 + 0 + 0 + 1344 + (-960) + 0 + 0 = 384$

Take the absolute value and multiply by $1/2$: Area $= \frac{1}{2}\cdot |384| = 192$

\ldots

\textbf{Final Answer} $\boxed{192}$ \xmark

\end{bluebox}
\end{minipage}\hfill
\begin{minipage}[t]{0.32\textwidth}
\begin{bluebox}[TaH Output]{width=\linewidth,  
height=0.95\textheight,  valign=top}

Okay\textcolor{blue}{,} let's try to tackle this geometry problem. It seems a bit \textcolor{blue}{ complex} with all these points and reflections\textcolor{blue}{,} \ldots
First, let me visualize triangle ABC. Points D, E, and B are on side AB with AD = 4, DE = 16, and EB = 8. \ldots
Let me set up a coordinate system\textcolor{blue}{. Let}'s place point A at the origin $(0, 0)$ \ldots Then, since $AB$ is a line segment, let \textcolor{blue}{ me} assume $AB$ is along the $x$-axis. So point $B$ would be at $(28, 0)$ \ldots
Now\textcolor{blue}{,} points $F$\textcolor{blue}{, G}, $C$ are on $AC$ \ldots Let \textcolor{blue}{me} define coordinates for $C$ as $(c_x, c_y)$ \ldots

\ldots

Compute the \textcolor{blue}{sum} of $x_i y_{i+1} - x_{i+1} y_i$ for each $i$ \ldots

Sum of all terms: $0 + \frac{1300 c_y}{91} + 0  -> \frac{52 c_y}{91} = \frac{(1300 - 52)c_y}{91} = \frac{1248 c_y}{91}$

Then, the area is $\frac{1}{2} \cdot \left|\frac{1248 c_y}{91}\right| = \frac{624 c_y}{91}$

\ldots

Compute the area of this heptagon \ldots 

Using the \textcolor{blue}{shoelace} formula \ldots

Term2: $-\!240$ \quad Term3: $840$ \quad Term4: $1176$ \quad Term5: $-\!840$ \quad Term6: $240$ \ldots

Sum: $0 - 240 + 840 + 1176 - 840 + 240 + 0 = 1176$

Take the absolute value and multiply by $1/2$:
Area $= \frac{1}{2}\cdot |1176| = 588$

\ldots

\textbf{Final Answer} $\boxed{588}$ \cmark

\end{bluebox}
\end{minipage}\hfill
\begin{minipage}[t]{0.32\textwidth}
\begin{bluebox}[AlwaysThink Output]{width=\linewidth, 
height=0.95\textheight, valign=top}

\begingroup
\color{blue}
Okay, so I need to find the area of the heptagon AFNBCEM. Let me first try to visualize the problem. There's triangle ABC. On side AB, there are points A, D, E, and B. The order is A, D, E, B. The distances given are AD=4, DE=16, and EB=8. So, starting at A, moving 4 units to D, then 16 units to E, then 8 units to B. So, the total length of AB is AD + DE + EB = 4 + 16 + 8 = 28 units. Got that.

\ldots

Points in order: D(4,0), E(20,0), F(13,h), G(65,h), back to D(4,0).

Compute the terms:

\ldots

Sum these terms: $0 + 20h - 52h - 4h = -36h$

Take the absolute value and multiply by 1/2: $Area = (1/2)*| -36h | = (1/2)*36|h| = 18|h|$

But the area is given as 288, so $18|h| = 288 \Rightarrow |h| = 16$.

\ldots

Now, apply the shoelace formula \ldots
Therefore, total sum is $8640/17$

Take the absolute value and multiply by 1/2: $Area = (1/2)*|8640/17| = 8640/34 = 4320/17 \approx 254.1176$

\ldots

Therefore, the answer is $4320/17$.

\endgroup
\textbf{Final Answer} $\boxed{\frac{4320}{117}}$ \xmark
\end{bluebox}
\end{minipage}
\end{figure*}




\end{document}